\documentclass[runningheads]{llncs}

\usepackage{eccv}

\usepackage{eccvabbrv}

\usepackage{graphicx}
\usepackage{booktabs}

\usepackage[accsupp]{axessibility}  

\usepackage{hyperref}

\usepackage{orcidlink}
\usepackage{makecell}
\usepackage{multirow}
\usepackage{booktabs}
\usepackage{bbding}
\usepackage{wrapfig}
\newcommand{\customfootnote}[1]{
    \begingroup
    \renewcommand{\thefootnote}{}
    \footnotetext{#1}
    \addtocounter{footnote}{-1}
    \endgroup
}

\begin{document}

\title{PEA-Diffusion: Parameter-Efficient Adapter with Knowledge Distillation in non-English Text-to-Image Generation} 

\titlerunning{PEA-diffusion}

\author{Jian Ma\orcidlink{0009-0004-0057-3033} \and
Chen Chen\orcidlink{0000-0003-3498-2527}$^{\dagger}$ \and
Qingsong Xie\orcidlink{0000-0001-6974-2019} \and
Haonan Lu\orcidlink{0000-0001-6332-2785}$^{\dagger}$}

\authorrunning{J.~Ma et al.}

\institute{OPPO AI Center \\
\email \{majian2,chenchen4,xieqingsong1,luhaonan\}@oppo.com}

\maketitle
\customfootnote{$^{\dagger}$Corresponding authors.}

\begin{abstract}
Text-to-image diffusion models are well known for their ability to generate realistic images based on textual prompts. However, the existing works have predominantly focused on English, lacking support for non-English text-to-image models. The most commonly used translation methods cannot solve the generation problem related to language culture, while training from scratch on a specific language dataset is prohibitively expensive. In this paper, we are inspired to propose a simple plug-and-play language transfer method based on knowledge distillation. All we need to do is train a lightweight MLP-like parameter-efficient adapter (PEA) with only 6M parameters under teacher knowledge distillation along with a small parallel data corpus. We are surprised to find that freezing the parameters of UNet can still achieve remarkable performance on the language-specific prompt evaluation set, demonstrating that PEA can stimulate the potential generation ability of the original UNet. Additionally, it closely approaches the performance of the English text-to-image model on a general prompt evaluation set. Furthermore, our adapter can be used as a plugin to achieve significant results in downstream tasks in cross-lingual text-to-image generation.
\end{abstract}

\section{Introduction}
\label{sec:intro}

Recent text-to-image (T2I) models, such as DALL-E2~\cite{ramesh2022hierarchical}, Imagen~\cite{saharia2022photorealistic}, and Stable Diffusion (SD)~\cite{rombach2022highresolution}, have opened up a new era of AI painting due to their ability to generate realistic, photo-level, and creative images. Numerous downstream tasks have further driven the iteration and update of this technology, as well as the activity of the community, including image editing, controllable generation, etc. However, mainstream text-to-image models only support English prompts, which means non-English speakers usually rely on translations to use them. This may cause translation errors and information loss due to culture-specific concepts. For example, one famous Chinese dish, ``Hongshao shizi tou'', may be translated to ``Braised lion head'' in English. Additionally, the introduction of large translation networks leads to high computation consumption.

To address this issue, recent researchers have created T2I generative models that process native languages directly, bypassing any need for translation. Taiyi-Bilingual~\cite{fengshenbang} incorporates a novel text encoder initialized with Chinese CLIP checkpoints, focusing only on training the text encoder. AltDiffusion~\cite{ye2023altdiffusion} introduces a multi-stage training process for multilingual T2I generation, which is relatively costly. Some experts begin training T2I models from scratch using different language data. For example, ERNIE-ViLG 2.0~\cite{ding2021cogview} is trained with native Chinese content and can generate high-quality images that reflect Chinese cultural elements. However, this approach is very expensive and often beyond the reach of many research facilities. Additionally, this method does not foster a community impact, necessitating the retraining of English models for various downstream or commonly used community tasks. GlueGen~\cite{qin2023gluegen} introduces a GlueNet module to align the input features of the original image generator by adding a new encoder. Despite its low training cost, its effectiveness in generating culturally relevant images in multiple languages and handling broad prompts is comparatively limited.

As such, exploring a parameter-efficient plug-and-play method that adapts the publicly available, English-naive, pretrained T2I models to suit non-English and culture-specific concepts is becoming more and more urgent. Consequently, we concentrate on a critical research question (RQ): Is it possible to devise a low-cost method to acquire a non-English T2I model by adapting the existing pretrained T2I models efficiently?
To this end, we take SD as the baseline, as it has shown surprisingly good results on T2I tasks. To finetune SD with minimal cost, we propose a novel parameter-efficient adapter (PEA) method with knowledge distillation (KD) to make full use of the pretrained SD model, referred to as PEA-Diffusion. To take advantage of the pre-trained SD model directly, the pre-trained SD is used as a teacher to guide student learning. Specifically, we propose the PEA module to align the representation space of the new condition encoder in the student with that of the teacher's image generator. PEA merely consists of a light-weight multi-layer perception (MLP), greatly simplifying the model construction process and preserving the image generation capabilities of the original SD model. Different from GlueNet~\cite{qin2023gluegen} which forces text feature alignment of two text encoders at encoder space, we propose KD loss to compel representation alignment at feature maps and logits space from the UNet module between our PEA-Diffusion and pretrained English SD models. We argue that our method boosts the image generator to understand the information from the new text encoder. To avoid catastrophic forgetting and reduce trainable parameters, UNet parameters are kept frozen during the entire training procedure.

Take Chinese SD as an example. We conduct a thorough experimental analysis on KD to help understand the effect of where to perform feature alignment between the teacher and the student. In addition, we provided intuitive analysis about what parameters should be updated, facilitating discovering the impact of the SD parameters for adapting SD into language-specific T2I models.
It should be noted that the method could be applicable for any other non-English T2I generation model.
Our core contributions are summarized as follows:
\begin{itemize}
\item  We propose a novel PEA-Diffusion model to resolve the T2I model’s English-native bias, supporting non-English prompts. The PEA aligns the feature and output spaces between the source and target UNet for bilingual prompts, leading to extremely low training costs.

\item PEA module is plug-and-play, it can be directly applied to various T2I tasks, including LoRA, ControlNet, Inpainting, compressed SD models that have been pruned and models accelerated through various strategies.

\item PEA-Diffusion can achieve remarkable image generation performance with culture-specific prompts compared with other cross-lingual or multilingual T2I models. It can also retain the general generation ability of the original English diffusion model with slight degradation on evaluation metrics.
\end{itemize}

\section{Related Work.}

\subsection{Multilingual Text-to-image Generation}
Diffusion models have shown a remarkable capability of yielding photorealistic and diverse images given textual descriptions, such as GLIDE~\cite{nichol2021glide}, Stable Diffusion~\cite{rombach2022high}, Imagen~\cite{saharia2022photorealistic}, DALL-E2~\cite{ramesh2022hierarchical}, and VQ-Diffusion~\cite{gu2022vector}. Despite their powerful generation abilities, they still only understand English input, leading to limitations. To support non-English
text prompt, Taiyi~\cite{zhang2022fengshenbang} finetunes the Chinese text encoder with a diffusion model. ERNIE-ViLG~\cite{zhang2021ernie} trains the T2I models from scratch.  AltDiffusion~\cite{ye2023altdiffusion} proposes a multilingual text
encoder to capture multiple language features, which serve as DM conditions. The model was trained in three stages: text encoder fine-tuning, $W_K, W_V$ fine-tuning, and full fine-tuning. GlueGen~\cite{qin2023gluegen} has designed a GlueNet module that aligns the original conditional input features of the Image Generator by introducing a new encoder. ENSAD~\cite{li2023translation} proposes a translation-enhanced multilingual T2I generation model by ensembling multilingual T2I and neural machine translation. BDM~\cite{liu2023bridge} proposes a backbone-branch network architecture by integrating an English T2I backbone with a Chinese native semantics injection branch. 
Despite their success in the non-English I2I generation, these works are either constrained to deal with distinctive language concepts or isolated from English-native T2I communities. At the same time, it is not possible to balance the lower cost of training with the performance of evaluation sets. IAP~\cite{hu2023efficient} proposes to treat images as pivots to establish the connection between the embedding space of different text encoders, but it doesn't consider to enhance cultural-related generation ability and expand to downstream tasks.

\subsection{Multilingual CLIP.}
CLIP uses contrastive language-image pretraining to connect text and image representations, based on massive English text-image sample pairs~\cite{radford2021learning}. Recently, some works have applied teacher learning~\cite{carlsson2022cross} or metric learning~\cite{aggarwal2020towards}  to train multilingual CLIP models, expanding CLIP to other languages. AltCLIP~\cite{chen2022altclip} takes this one step further enhancing the language capabilities, which is implemented by teacher learning and contrastive learning by leveraging a pretrained multilingual text encoder, XLMR~\cite{conneau2019unsupervised}. mCLIP~\cite{chen2023mclip} achieves multilingual CLIP by aligning the CLIP text model and a multilingual text encoder. To effectively transfer CLIP to language-specific scenarios, some researchers train language-specific CLIPs on large-scale native image-text pairs, e.g., Chinese CLIP~\cite{yang2022chinese}.
In order to promote PEA-Diffusion to effectively understand language-specific concepts, we append the PEA module after the language-specific CLIP or Multilingual CLIP and train it to uncover non-English cultural concepts within the T2I. The goal is to establish a bridge between the new CLIP and T2I core structures, such as UNet using the PEA module.

\subsubsection{Knowledge Distillation.}
KD~\cite{hinton2015distilling} transfers the knowledge in the output logits of a teacher model with a large size to a smaller student model without obvious performance degradation. The feature map in the hidden states and attention output space can also be used for KD to help students mimic teacher capability~\cite{jiao2019tinybert,hou2020dynabert}. Masked Generative Distillation~\cite{yang2022masked} introduces a feature masking scheme into KD to improve students’ representation power by forcing them to recover the masked features. Inspired by masked image modeling, the supervised masked KD model is proposed by inserting label information into self-distillation frameworks. However, to the best of our knowledge, KD has not been fully explored for training multilingual T2I generations. In this paper, we apply KD methods to language transfer in T2I and further harness the power of the PEA module through combination to achieve significant improvements.

\section{Method}
A naive approach for generating images from non-English text is to substitute the CLIP text encoder in SD with a language-specific CLIP encoder. The text encoder is a crucial part of T2I models since exact and detailed text embedding is necessary for accurate generation. However, altering an existing text model to a different language-specific one is difficult. This difficulty arises because SD's text encoder and image generator are closely integrated, making it ineffective to just swap out the condition encoder. The primary issue is the mismatch between the new text encoder and the old image generator. Furthermore, our experiments show that joint fine-tuning depends heavily on high-quality text-image pairs and often fails due to catastrophic forgetting when updating well-trained parameters. In light of these insights, we keep the pretrained model's parameters fixed and develop a lightweight adapter with KD to synchronize the new text encoder and the image generator.

\begin{figure}[tb]
	\centering
	\includegraphics[width=0.8\textwidth]{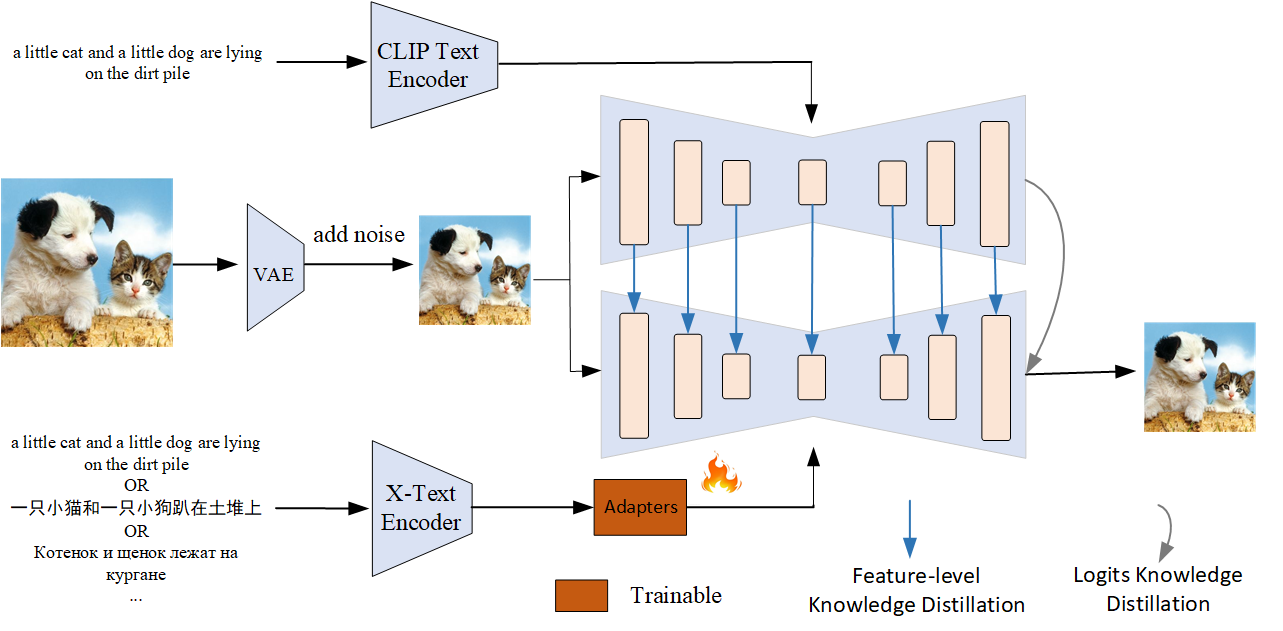}
 \caption{\small{An overview of the proposed PEA-Diffusion. Notice that only the lightweight adapter is trainable through the whole training process.}}
\label{framework}
\end{figure}

Figure~\ref{framework} illustrates the overall architecture of PEA-Diffusion for cross-lingual transfer in T2I generation. To inherit the knowledge of the pretrained diffusion model, we use the pretrained SD as a teacher to supervise PEA-Diffusion learning through KD. For language-specific T2I generation, we use the corresponding language-specific CLIP as a text encoder to capture text information. The generator in PEA-Diffusion is the same UNet as that used in SD. A light-weight MLP adapter is added on top of the CLIP text encoder in PEA-Diffusion, facilitating knowledge transfer from teacher to student.
 
\subsection{Preliminary} 
The diffusion model is composed of a forward diffusion process and a reverse denoising process. The forward process gradually adds random noise to clean data and diffuses it into pure Gaussian noise, while the reverse process reverses the diffusion process to create satisfactory samples from the Gaussian noise\cite{ho2020denoising}.

Specifically, for an input image $x_0 \in \mathbb{R}^{H \times W \times 3}$, the encoder $\mathcal{E}$ of the auto-encoder transforms it into a latent representation $z_0 \in \mathbb{R}^{h \times w \times c}$, where $f=H/h=W/w$ is the downsampling factor and $c$ is the latent feature dimension.
The diffusion process is then performed in the latent space, where a conditional UNet~\cite{ronneberger2015u} denoiser $\epsilon_\theta$ is employed to predict noise $\epsilon$ with current timestep $t$, noisy latent $z_t$ and generation condition $C$. In T2I scenarios, the condition $C=\tau_\theta(y)$ is produced by encoding the text prompts $y$ with a pretrained CLIP~\cite{radford2021learning} text encoder $\tau_\theta$. Therefore, 
the overall training objective of SD is defined as
\begin{equation}
    \label{sd}
    \mathcal{L}_\text{SD} = \mathbb{E}_{\mathcal{E}(x_0),C,\epsilon \sim \mathcal{N}(0,1),t} \Big[ \lVert \epsilon - \epsilon_\theta (z_t, t, C)\rVert_2^2 \Big]
\end{equation}

\subsection{Cross-Lingual Transfer} 
\textbf{Objective}. Given sets of parallel corpus $Y^T=\{y_i^t\}_{i=1}^N$ and $Y^S=\{y_i^t\}_{i=1}^N$ and ${Y^S}^{'}=\{y_i^t\}_{i=1}^M$ and the corresponding images $I=\{img\}_{i=1}^N$ with $N$ denoting the total number of paired samples, $Y^T$ represents the English corpus that will be input into the teacher model, $Y^S$ represents the non-English corpus with culturally irrelevant content, and ${Y^S}^{'}$ represents the non-English corpus with cultural relevance, they will be input into the student model.
The teacher unet denoiser $\epsilon^T$ projects pairs $(Y^T, I)$ into source middle feature maps and source logits output. 
Likewise, the student unet denoiser $\epsilon^S$ projects pairs $(Y^S, I)$ into target middle  feature maps and target logits output. Intuitively, there exists severe distribution mismatch between the source and target output, as well as between $\epsilon^S$ and $\epsilon^T$ due to the different languages. 
Therefore, our target is to make $\epsilon^S$ match with $\epsilon^T$, so as to transfer teacher knowledge to the student, enabling cross-lingual T2I generation. For the student, we use a language-specific CLIP text encoder instead of the original English text encoder to enhance token- and sentence-level lingual-native representations.
For brevity, we omit the subscripts of $\mathbb{E}_{\mathcal{E}(x_0),\tau_\theta(y),\epsilon \sim \mathcal{N}(0,1),t}$  and $z_t,t$ conditions in the following notations.

\textbf{Knowledge Distillation for T2I generation}.
The original KD leverages Kullback-Leibler (KL) divergence to measures the difference between the categorical probability
distribution of the teacher's logits output and the student's logits output~\cite{hinton2015distilling}. However, in the task of T2I generation, since the middle feature maps are the latent value of pixels instead of categorical
probability distribution, KL divergence is unable
to measure the difference between student and teacher.
Consequently, we replace KL divergence with the L2-norm distance between the logits output from student and teacher to extend raw KD for T2I generation, whose loss function is formulated as:
\begin{equation}
  \mathcal L_{LKD}=\mathbb{E}\Big[\lVert \epsilon^T(y^T) - \epsilon^S(y^S)\lVert_2^2 \Big]
\end{equation}
where $y^T \in Y^T$ and $y^S \in Y^S$. $\mathcal L_{LKD}$ merely enforces student distribution to match teacher distribution at logits output layer.
To further reduce distribution shift of teacher and student, we boost feature alignment of the middle feature maps between $\epsilon^S$ and $\epsilon^T$:
\begin{equation}
    \mathcal L_{FLKD} = \mathbb{E}\Big[\sum_{l}\lVert f(g_l^T(y^T)) - f(g_l^S(y^S))
    \lVert_2^2 \Big]
\end{equation}
where $f$ represents the norm operation, $g_l^T$ and $g_l^S$ represent the feature maps of the $l$-th layer in a predefined set of unet layers from the teacher and the student, respectively. With respective weights $\lambda_{FL}$ and $\lambda_{L}$, the final KD objective is formalized as below:
\begin{equation}
    \mathcal L_{KD}=\lambda_{FL}\mathcal L_{FLKD}+\lambda_{L}\mathcal L_{LKD}
\end{equation}

\textbf{Adapter}. Due to the dimension difference between English CLIP and non-English CLIP, we propose an adapter for simple feature transformation to align feature dimension. Moreover, the adapter is trained to learn language-specific information. The adapter is implemented by a MLP with only 6M parameters. To accelerate training and avoid destroying the generalization capability of the pretrained models, we freeze the CLIP encoder and image generator. As a result, the proposed PEA-Diffusion is both parameter- and
computation-efficient because just the adapter is trained.
\subsection{Training Strategy and Final Objective}
For our goal of training a multi-language T2I generation model, we only need to train on English data since both the new CLIP text encoder and the teacher model's CLIP text encoder can accept English inputs. In this case, we can use aligned KD loss $\mathcal L_{KD}$. For our goal of training a language-specific T2I generation model, we adopt a hybrid training strategy, for the target language's parallel corpus, we only need to use KD loss $\mathcal L_{KD}$ to align the distribution. On the other hand, we also have language-specific culture-related data ${Y^S}^{'}$. Because there is a significant translation error when translating these data into English and the teacher model cannot provide high-quality knowledge guidance for this type of data, we only use $\mathcal{L}_\text{SD}$ \eqref{sd} for improving culture understanding ability.
The final objective is formalized as follows:
$$\mathcal L = \begin{cases}
    \lambda_{FL}\mathbb{E}\Big[\lVert \epsilon^T(y^T) - \epsilon^S(y)\lVert_2^2 \Big] + \lambda_{L}\mathbb{E}\Big[\sum_{l}\lVert f(g_l^T(y^T)) - f(g_l^S(y))
    \lVert_2^2 \Big], & y \in Y^S \\
\mathbb{E} \Big[ \lVert \epsilon - \epsilon^S (y)\rVert_2^2 \Big], & y \in {Y^S}^{'} \\
\end{cases}$$

\section{Experiments}
\label{sec:blind}

\begin{figure*}[tb]
	\centering
	\includegraphics[width=1\textwidth]{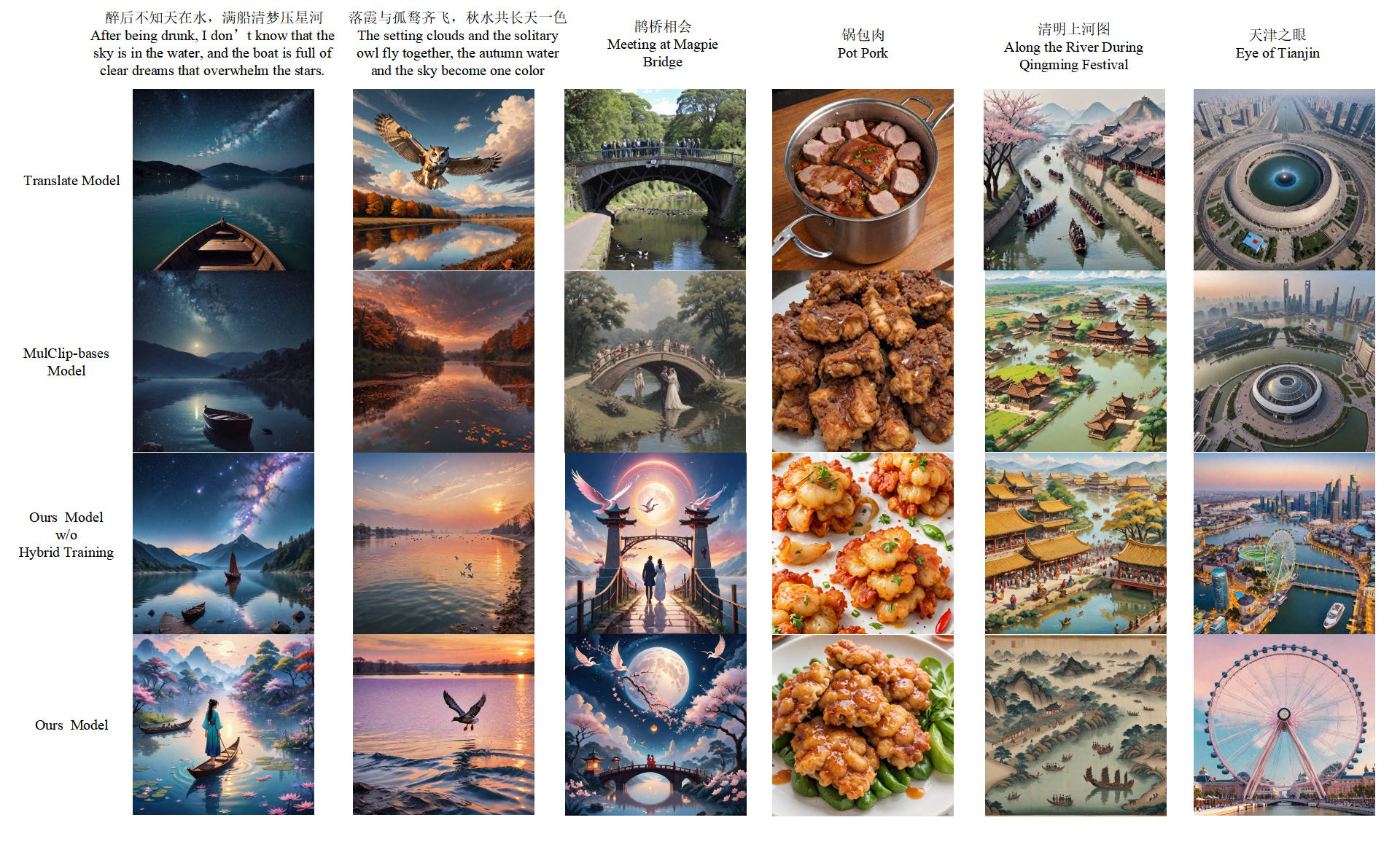}
 \caption{\small{Image generation visualization of different models in Chinese specific language.}}
\label{fig_case}
\end{figure*}

\subsection{Data Preparation}
We download the Laion2B-multi~\cite{schuhmann2022laion5b} dataset along with English translations for image descriptions. Based on the open-source CLIP models available for different languages: Chinese CLIP, Russian CLIP, Japanese CLIP, Korean CLIP, and Italian CLIP, we determine to conduct experiments on the above five languages. For Chinese, we also collect WuKong~\cite{gu2022wukong} data with Chinese cultural image-text pairs and Chinese-English parallel corpus translations of Laion2B-en. For more data, model, and code details, please refer to the Appendix. 

\subsection{Implementation Details and Evaluation}

\textbf{Implementation details}.
We conduct all experiments on Stable Diffusion (SD)-~\cite{rombach2022high},and Stable Diffusion XL (SDXL)-~\cite{podell2023sdxl} like backbones.
Take SDXL base model as an example. PEA-Diffusion consists of VAE, UNet, CLIP text encoder, and the adapter module, comprising 3 billion parameters, out of which a mere 6 million parameters ($\approx$ 0.2\%) are trainable.
We set $\lambda_{FL}$ 0.1 and $\lambda_{L}$ 1 respectively based on the loss performance.
The entire model is trained on 64 A100 GPUs for 70000 steps with a batch size of 5 per GPU.

\textbf{Test benchmark}. To evaluate the capability of PEA-Diffusion to generate images and capture culture-specific concepts of different languages, we introduce two datasets: Multilingual-General (MG) for generation quality evaluation and Multilingual-Cultural (MC) for culture-specific concepts evaluation. We collect 200 prompts for MG, which covered content such as people, animals, landscapes, and fantasy themes. The descriptions include diverse styles, such as cartoon, anime, illustration, watercolor, and cyberpunk. The data comes from various major art websites where users provide real prompts. On the other hand, we also construct MC data including 200 prompts. The collection is mainly based on the ChatGPT tool and consists of culturally relevant content for each specific language, including but not limited to local cuisine, architecture, clothing, and distinctive cultural features. 

\textbf{Evaluation metrics}.
For the general evaluation data MG, the widely used FID metric can only measure the feature distance between generated images and specific images. It often fails to provide accurate assessment for high-quality generated images, as emphasized and experimentally validated by SDXL~\cite{podell2023sdxl} and other studies. Therefore, in order to evaluate the quality of generated images more accurately, we introduce two human preference models as our evaluation metrics: ImageReward~\cite{xu2023ImageReward} and PickScore~\cite{kirstain2023pickapic}. 
Additionally, we continue to utilize OpenCLIP ViT-bigG to measure the similarity between images and text.
As for the language-specific evaluation data MC, constructed for each language, to ensure a fair comparison of content relevance, we only consider the CLIP model trained in the specific language to calculate the correlation between generated images and their corresponding text. Since the general CLIP model can only process non-English data after translation into English, this process can introduce significant performance degradation. Furthermore, our paper's focus is precisely on addressing the errors caused by translating non-English prompts into English to generate images. Therefore, other metrics that involve translation processes, such as the two models based on human preference mentioned above, will no longer be valid for evaluation.

\textbf{Baseline methods.}  We generated image results in specific languages based on two T2I base models: SD1.5 and SDXL. We choose the following four categories of approaches as our baselines.
\begin{itemize}
\item Translation: we apply Google translation API.

\item AltDiffusion~\cite{ye2023altdiffusion}: we test two multilingual SD models based on the open-source AltDiffusion framework: AltDiffusion-m9 and AltDiffusion-18.

\item GlueGen~\cite{qin2023gluegen}: We evaluate our model on three languages that overlap with the checkpoint provided by GlueGen, which are Chinese, Italian, and Japanese. For the other two languages, Russian and Korean, we implement GlueGen based on its code and our data. In addition, to further compare the impact of different language-specific CLIPs on the results, we also retrain the GlueGen based on five languages' CLIP text encoders. 

\item Fine-tuning with LoRA~\cite{hu2021lora} and $W_K, W_V$ (key and value projection matrices): As low-cost experiments for comparison, we also fine-tuning the model to transfer language capabilities.



\end{itemize}

\subsection{Experimental Results}

\textbf{Generation analysis.} We validate our approach based on two different T2I base models.
Firstly, to ensure a fair comparison with GlueGen and AltDiffusion, we train PEA-Diffusion based on the SD1.5 model.
Secondly, to validate our low-cost experiment approach, we train a larger model based on SDXL. The compared methods are LoRA fine-tuning with rank 64 and $W_K, W_V$ fine-tuning, both of which are parameter-efficient fine-tuning strategies. For the second approach, as proposed in~\cite{kumari2023multi}, when updating the mapping from given text to image distribution, only updating $W_K^{(i)}, W_V^{(i)}$ in each cross-attention block $i$ is sufficient since text features are the only input to the key and value projection matrix.
Furthermore, to validate the wide application of the proposed approach with different text encoders, we train our model with a multi-language T5 encoder and a multi-language CLIP encoder. This type of training only requires English text data, which makes it even simpler and more efficient. Fig.~\ref{fig_case} illustrates the comparisons between translation and our methods with different training strategies. It can be observed that our PEA-Diffusion can achieve Chinese culturally related image generation, but translation totally fails in these cases.
\begin{figure*}[ht]
\centering
\captionof{table}{\small{CLIPScore with different specific languages. For Chinese, we evaluated two general evaluation metrics with MG data. $\dagger$ Indicates open source models.}}
\label{tab:comparisons}
\resizebox{1\linewidth}{!}{
\centering
\begin{tabular}{ccccccccc}
\toprule
& & \textbf{zh-MC} &\multicolumn{2}{c}{\textbf{zh-MG}} & \textbf{ru-MC}   & \textbf{it-MC}& \textbf{ko-MC} & \textbf{ja-MC} \\ 
\cmidrule(l){4-5} 
\textbf{Model Base} & \textbf{Methods} & \textbf{CLIPScore\_zh} & \textbf{ImageReward} &\textbf{ PickScore} &\textbf{CLIPScore\_ru} &\textbf{CLIPScore\_it} &\textbf{CLIPScore\_ko} &\textbf{CLIPScore\_ja} \\\midrule

\multirow{6}{*}{\textbf{SD1.5}} & Translate  &  0.1686 & 0.2427 & \textbf{0.1799} & 0.4911 & 0.3556 & 0.1904  & 0.2885   \\
 & Altdiffusion m9~\cite{ye2023altdiffusion}$\dagger$  & 0.2100 & -0.6966 & 0.1202 & 0.3715 & 0.3177 & 0.1906  & 0.2461 \\
 & Altdiffusion m18~\cite{ye2023altdiffusion}$\dagger$  & 0.2172 & \textbf{0.3316} & 0.1648 & 0.4730 & 0.3502  & \textbf{0.1917}  & \textbf{0.3008} \\
 & GlueGen~\cite{qin2023gluegen}$\dagger$ & 0.1700 & -0.5187 & 0.1224 & 0.4184  & 0.3065 & 0.1769  & 0.2506  \\
 & GlueGen~\cite{qin2023gluegen}   & 0.2153 & -0.4685 & 0.1256  & 0.3987  & 0.2928  &  0.1843  & 0.2465  \\
 & Ours & \textbf{0.2524} & 0.0985 & 0.1701 & \textbf{0.5007} &  \textbf{0.3728} & 0.1896  & 0.2501 \\\midrule

\multirow{5}{*}{\textbf{SDXL}} & Translate   & 0.2383 & \textbf{1.3647} & \textbf{0.2402}  & 0.4844 & 0.3591   & 0.2033  & 0.2958\\
 & LoRA   & 0.2366 & 0.1217 & 0.1598  & 0.4656 &  0.3476  &  0.2034  & 0.1940  \\
 & $W_k$ and $W_v$  & 0.2485 & 0.1413 & 0.1621  &  0.4785 & 0.3495  & 0.2088  &  0.2022 \\
  & Ours w/ MulT5 & 0.2301 & 0.8254 & 0.2001 & 0.4568  & 0.3805 & 0.1908  & 0.2996  \\
 & Ours w/ Multilingual-CLIP & 0.2407 & 1.2122 & 0.2324 & 0.4673  & 0.3856 & 0.2157  & \textbf{0.3007} \\
 & Ours & \textbf{0.2610} & 1.0245 & 0.2244 & \textbf{0.5012}  & \textbf{0.4191} & \textbf{0.2194}  &  0.2544 \\\bottomrule

\end{tabular}}
\end{figure*}

As shown in the comprehensive results of Table~\ref{tab:comparisons}, when compared with the translation method, our PEA-Diffusion outperforms it for almost all languages (except Japanese) in terms of CLIPScore for MC datasets. It is obvious that the translation method cannot capture the essence of different languages' cultures, leading to a semantic gap between the original prompts and the translated English prompts. However, our approach can use additional culturally related image-text pairs to train the adapter, which can align the language understanding ability of the text encoder with the generation ability of the UNet. Even though the UNet is frozen during the whole training process, we argue that our PEA-Diffusion can still stimulate the potential generation ability for cultural-related images of the original UNet. Since SD1.5 and SDXL are all trained on billions of image-text pairs that contain wide image domains, our approach can successfully awaken their generative abilities in different cultural contexts. In terms of ImageReward and PickScore for the MG dataset, we must admit that PEA-Diffusion is slightly inferior to translation methods. This is because using other text encoders (e.g., Chinese CLIP and Multilingual CLIP) cannot ensure complete alignment between the text embedding and the UNet feature spaces.

Even though Altdiffusion retrains the multilingual CLIP text encoder and UNet on billions of image-text pairs, it still performs worse than our approach in terms of almost all metrics for all languages. GlueGen is the most related work on our concerned topic. The statistics presented in Table~\ref{tab:comparisons} reveal that both checkpoints provided by the authors or re-implemented by ourselves are significantly inferior to our PEA-Diffusion, especially for the ImageReward and PickScore metrics, resulting in very low generation quality since only aligning distributions of text embedding for different languages is very hard, even with its sophisticatedly designed GlenNet. In addition, it cannot generate culturally related images. Therefore, it's worth mentioning that our core contribution is that we align the model's generation ability through UNet distillation only by a MLP-like adapter between different languages. We don't forcibly align the text embedding spaces of different languages.

Additionally, we undertake various parameter-efficient fine-tuning strategies for comparison. The fine-tuning techniques, including LoRA and $W_k$ and $W_v$, show subpar performance based on ImageReward, PickScore, and CLIPScore metrics. This is attributed to the fact that these fine-tuning methods depend significantly on high-quality image-text datasets, whereas our PEA-Diffusion requires merely some parallel multilingual texts and images and lacks stringent quality requirements. Further experiments can be found in the Appendix. 

\textbf{Analysis on different languages.}
The data in Table~\ref{tab:comparisons} indicate that the CLIPScores for Russian, Italian, and Korean either match or exceed those of other baseline methods by a notable margin. For Japanese, however, the CLIPScore is slightly lower compared to translation and AltDiffusion. We ascribe this to the CLIP model's limited encoding ability for Japanese, which fails to align with the SD model via the adapter layer. Additionally, with only 1.6M training samples for Korean, our method still achieves effective language transfer, underscoring that large-scale training datasets are not essential. Further visualization examples of culturally-related prompts can be found in the Appendix. 

\begin{wrapfigure}{r}{5cm}
\centering
\includegraphics[width=0.4\textwidth]{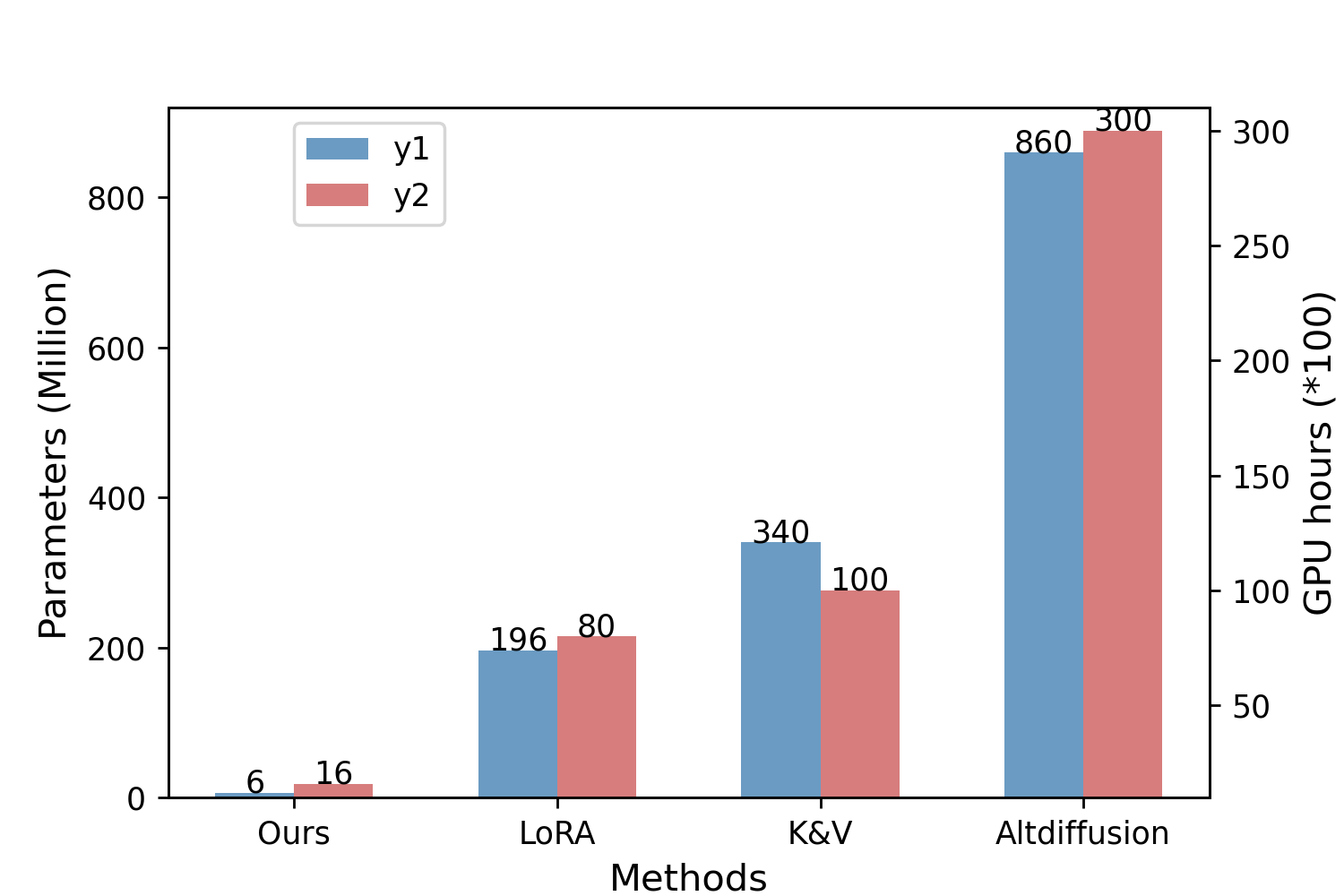}
\caption{Training parameters(M) and cost comparison for different methods.}
\label{cost_col}
\end{wrapfigure}
\textbf{Analysis of training cost.}
Our PEA-Diffusion freezes the CLIP text encoder, the UNet, and VAE during the whole training process, only the MLP mapping adapter is trained, accounting for merely 6M training parameters, which is 0.2\% of total parameters for SDXL-based models. We employed 64 A100 GPUs to train 70000 steps, resulting in 1600 GPU hours. As a comparison, AltDiffusion~\cite{ye2023altdiffusion} trained the model for three stages with nearly 30000 GPU hours, which is $\approx 20$ times our training cost. In addition, our method achieves higher performance in terms of almost all metrics and the five experimental languages. We also re-implemented and trained GlueGen~\cite{qin2023gluegen} with the same dataset constructed by ourselves. It indeed achieves very low training costs with only 1 A100 GPU for 24 hours, but results in very poor performance no matter for image quality (-0.5 ImageReward score) or language transferring ability. Finally, we also conduct other parameter-efficient fine-tuning strategies (e.g., LoRA and tuning only $W_k$ and $W_v$ matrices of attention layers) and achieve much lower performance and higher training costs, which demonstrated the effectiveness and efficiency of our PEA-Diffusion.
Figure~\ref{cost_col} only shows the relationship between training parameters and GPU hours of different methods. For more information on model fitting and training data, as well as adapter module parameter variations, please refer to Appendix. 


\subsection{Ablation Studies}

Since Chinese data has certain advantages in terms of data quality and quantity, to eliminate errors caused by data, our ablation experiments are all based on Chinese data. The experiments include four parts:
First, we conduct a baseline experiment where we only perform standard model fine-tuning without incorporating any KD strategies.
Secondly, inspired by GlueGen~\cite{qin2023gluegen}, we employ text encoder KD to explore the impact of distillation at different positions in the SD framework.
Next, we perform distillation based on the UNet model, including feature distillation between different UNet layers and logits distillation for the final output.
Finally, we combine all of the distillation positions described above.

\begin{figure}[ht]
\centering
\captionof{table}{\small{Ablation results for different knowledge strategies.}}
\label{Ablation}
\resizebox{0.8\linewidth}{!}{
\begin{tabular}{ccccc}
\toprule
\textbf{Evaluation} & \textbf{ImageReward} & \textbf{PickScore} & \textbf{CLIPScore\_en} & \textbf{CLIPScore\_zh} \\\midrule
Fine-tuning & -1.3154 & 0.1854 & 0.3278 & 0.1901  \\
Fine-tuning w/ TKD & -0.9875 & 0.1997 & 0.3506 & 0.2287 \\
Fine-tuning w/ ULKD & -1.0447 & 0.1957 & 0.3331 & 0.1968  \\
Fine-tuning w/ UFKD & 0.9556 & 0.2174  & 0.4085 & 0.2557  \\
Fine-tuning w/ UTKD & 0.8490 & 0.2113  & 0.4011 & 0.2461 \\\midrule
Ours(Fine-tuning w/ UKD) & \textbf{1.0245} & \textbf{0.2244} & \textbf{0.4224} & \textbf{0.2610}  \\\bottomrule
\end{tabular}}
\end{figure}



\textbf{Effectiveness of KD.}
First, KD is crucial for guiding models onto the right learning trajectory for rapid convergence and is vital for the PEA-Diffusion method. As indicated in the first row of Table~\ref{Ablation}, the standalone fine-tuning method yields notably inferior performance in both image quality (ImageReward and PickScore) and image-text alignment (CLIPScores).

\textbf{Effectiveness of UNet Feature KD (UFKD).}
Our KD method consists of two parts: the first part involves distilling the logits results generated by the UNet (ULKD), we then introduce a distillation of intermediate layer features in the UNet, which serves as the primary knowledge guidance mechanism. As shown in the third and fourth lines of Table~\ref{Ablation}, we can observe that logit distillation can only improve marginally when compared with direct fine-tuning whereas feature distillation boosts the performance remarkably. We attribute this phenomenon to the generation ability of the large amount of UNet parameters, so that intermediate feature supervision is essential for knowledge transfer. 

\textbf{Effectiveness of UNet KD.}
Furthermore, different distillation positions have different effects. Our initial focus is on aligning the feature spaces of different text encoders, so a natural idea is to distill only the feature information of the teacher model's text encoder (TKD). However, as the results show in the second line of Table~\ref{Ablation}, although the overall performance exceeds the pure fine-tuning method, it is still far from our best performing approach. This illustrates that simply aligning the text encoder is insufficient and ineffective. This statistic also matches the experimental results of GlueGen in Table~\ref{tab:comparisons}: simply aligning text embedding spaces cannot obtain satisfactory image generation results. Therefore, this further proves the necessity of distilling the features of the UNet itself.

Finally, we perform KD on both the UNet and text encoder (UTKD). However, the fifth line of Table~\ref{Ablation} shows that its effect is worse than performing only UNet distillation. We suspect that the addition of text encoder KD actually weakens the learning of the SD core UNet structure. The main knowledge ability for generating images is mainly influenced by the UNet model, which is responsible for the adapter alignment feature in PEA-Diffusion. More ablation experiments can refer to Appendix. 

\subsection{Exploring the Domain Adaptation of PEA-Diffusion}

To demonstrate the domain adaptation of our model with minimal data guidance, we first constructed a dataset of 20,000 parallel image and text pairs in the domain of human. We then trained the models based on the models fine-tuned $W_K^{(i)}, W_V^{(i)}$ mentioned in section 4.3 with UKD, and evaluated on the MG and MC dataset. The results on Chinese are shown Table \ref{Domain}.
\begin{figure}[tb!]
\centering
\captionof{table}{\small{Domain adaptation with 20,000 data in Chinese specific language.}}
\label{Domain}
\resizebox{0.8\linewidth}{!}{
\begin{tabular}{ccccc}
\toprule
\textbf{Evaluation} & \textbf{ImageReward} & \textbf{PickScore} & \textbf{CLIPScore\_en} & \textbf{CLIPScore\_zh} \\\midrule
$W_k$ and $W_v$  & 0.1413 & 0.1621  & 0.3674 & 0.2485 \\\midrule
Ours(Fine-tuning w/ UKD) & \textbf{0.8368} & \textbf{0.2145} & \textbf{0.4098} & \textbf{0.2501}  \\\bottomrule
\end{tabular}}
\end{figure}

We observed significant improvements in the three metrics on the MG test data. This suggests that PEA can be guided by a small amount of data to perform out-of-domain knowledge transfer.

\subsection{Adaptability Experiment of English Downstream Model}

\begin{figure*}[!h]
	\centering
	\includegraphics[width=1\textwidth]{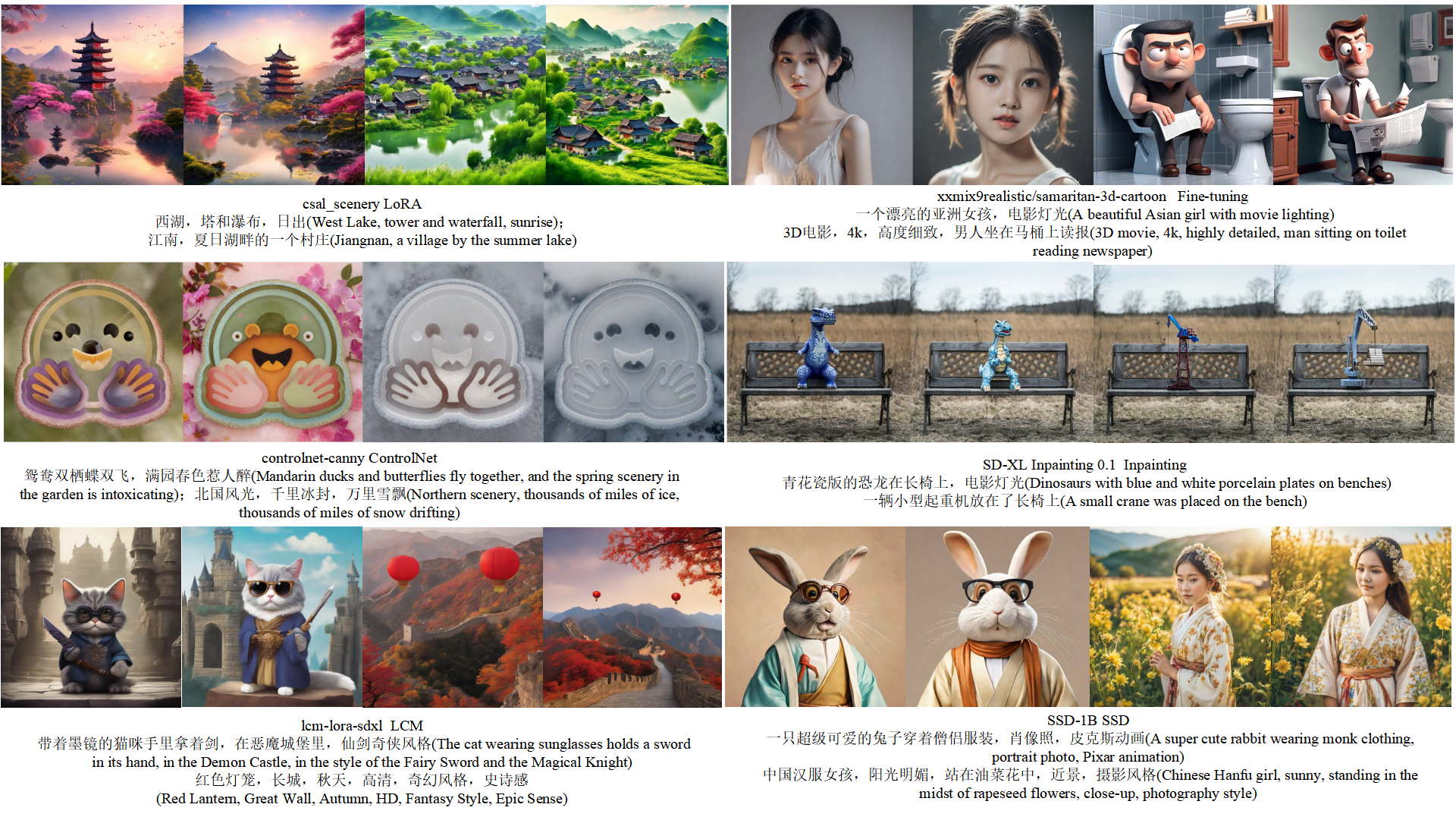}
 \caption{\small{Downstream task experiments of our proposed PEA. By only adding the trained adapter between the Chinese CLIP text encoder and the UNet, it can achieve language transfer for fine-tuning, LoRA, ControlNet, Inpainting, LCM and compressed models.}}
\label{fig_downstream0}
\end{figure*}

In this section, we would like to emphasize the plug-and-play feature of our PEA module. Currently, a wide range of community resources, such as fine-tuned models, LoRA, Dreambooth, ControlNet, compressed SD models, and accelerated sampling SD models, are derived from the English SD models, so that the English community is much more mature than other languages. Therefore, one key challenge we must overcome is bridging the compatibility gap between the new model and the English SD community after language transfer. Since our approach uses feature-level and logit-level KD to train only the PEA module while keeping the core UNet parameters unchanged, it naturally maintains compatibility with the English SD community and can seamlessly integrate with various English SD community plugins.
Interestingly, the PEA module remains plug-and-play for models that have undergone changes in base SD UNet parameters in contrast to PEA-Diffusion but whose overall structure remains similar. This includes models such as fine-tuned SD and compressed SD trained using pruning and distillation techniques, among others. In the following, we will introduce several downstream image generation applications based on the SD or SDXL foundation model.

\textbf{Fine-tuned checkpoint.}
We use two fine-tuned SDXL-based model checkpoints: ``xxmix9realistic'' and ``samaritan-3d-cartoon''. The ``xxmix9realistic'' model is trained on person, photorealistic, sexy, female, woman, girls, etc. The ``samaritan-3d-cartoon'' model is trained on anime, Disney, 3D, Pixar, etc.

\textbf{LoRA.}
``csal\_scenery'' is a SDXL-based LoRA model that contains common ancient Chinese architectural forms such as pavilions, pagodas, towers, gardens, courtyards, bridges, temples, palaces, towns, etc.

\textbf{ControlNet.}
``controlnet-canny'' is trained on SDXL under canny condition.

\textbf{Inpainting.}
SDXL Inpainting 0.1 is a latent T2I diffusion model capable of generating photorealistic images given any text input, with the extra capability of Inpainting the pictures by using a mask.

\textbf{LCM.}
Latent Consistency Models~\cite{luo2023latent} have achieved impressive performance in accelerating T2I generative tasks, producing high-quality images with minimal inference steps. ``lcm-lora-sdxl'' is an acceleration module based on SDXL.

\textbf{SSD.}
The Segmind Stable Diffusion Model is a condensed version of the SDXL that is 50\% smaller in size, delivering a 60\% increase in speed while preserving high-quality text-to-image (T2I) generation capabilities.

In Figure~\ref{fig_downstream0}, we have shown that the PEA module can be used as a plugin between the Chinese CLIP text encoder and the UNet, achieving the ability to transfer language from English to Chinese directly.
For more details of the above-mentioned checkpoints and generated images, please refer to our Appendix. 

\section{Conclusion and Limitation}
\textbf{Conclusion.} The high training expenses and limited training data available for multilingual T2I models have driven us to search for a more lightweight solution. In this paper, we introduce a straightforward and efficient method using KD and adapters. By focusing solely on training the adapter and keeping other parameters fixed, our approach harnesses the capability of T2I models for multilingual cultural comprehension and generation. Our method not only achieves significantly better results for culturally relevant data but also surpasses existing models while maintaining general synthesis capability. Comprehensive experiments demonstrate that our method is cost-effective, versatile, and easy to integrate.

\textbf{Limitation.} First, the effectiveness of the CLIP text encoder significantly influences both the quality of image generation and text relevance. Utilizing a weaker language-specific CLIP could impact the performance of PEA-Diffusion in broad open domains and may slightly compromise the model's generalization abilities. Second, the upper limit of our method is constrained by the capabilities of the English base model, indicating that our approach cannot exceed its inherent potential. Moving forward, we aim to conduct additional research to mitigate these limitations.

\clearpage  

%
%
\bibliographystyle{splncs04}
\bibliography{main}

\clearpage
\setcounter{page}{1}

\begin{center}
    \huge
    \textbf{Appendix}
    \\[16pt]
\end{center}

Appendix~\ref{data} provide information on how we organize and prepare the training data. We also present URL links to all publicly available models and code used in the paper.

Appendix~\ref{Multilingual} analyze the performance of the Chinese language model with different training data resolutions and present detailed quantitative analysis of all languages in both the general evaluation set (MG) and the culture-specific evaluation set (MC) as the number of training samples increases. Finally, we show qualitative analysis in other languages.

Appendix~\ref{Hybrid} provide the results of the ablation experiments on hybrid training and details of ablation experiments in Sec. 4.4.

Appendix~\ref{Convergence} analyze low-cost experiments.

Appendix~\ref{LLM_suppl} analyze the performance of LLMs as text encoder for T2I.

Appendix~\ref{table1} analyze the qualitative results for Sec. 4.3 Table 1 in the general domain.

Appendix~\ref{user_study} analyze the user study.

Appendix~\ref{Adaptability} provide additional visualizations of down-stream text-to-image generation tasks in Chinese.

\section{Data, Model and Code}
\label{data}
\subsection{Data Statistics}
Based on the SDXL~\cite{podell2023sdxl}, we employ a multi-scale training strategy using ratio bucketing. Based on the analysis of data resolutions and quantities for different languages, we observe limited data availability around the resolution of 1024. Hence, we chose to focus on the data at around 640 resolution for training. However, we perform additional training at a resolution of around 1024 specifically for the Chinese data, considering its sufficient quantity.
Based on the SD1.5~\cite{rombach2022high}, we uniformly resize and crop the data to 512 resolution data for training. Table~\ref{data_tab} details the training data amount at different resolutions of various languages. 

\begin{figure}[!ht]
\centering
\setlength{\belowcaptionskip}{0.5em}
\setlength{\abovecaptionskip}{0.5em}
\captionof{table}{\small{Total amount of data with different resolutions.}}
\label{data_tab}
\resizebox{1\linewidth}{!}{
\begin{tabular}{ccccc}
\toprule
\textbf{Languages} & \textbf{256 × 256 pix} & \textbf{512 × 512 pix} & \textbf{1024 × 1024 pix} \\\midrule
zh & 500M & 160M &  51M   \\
ru & 130M & 51M & 5M  \\
ja & 64M & 18M &  2.8M \\
it & 44M & 18M & 2.6M  \\
ko & 8.5M & 1.6M & 0.3M \\\bottomrule
\end{tabular}}
\end{figure}

\subsection{Model or Code Path}
The datasets, models and codes used in this paper are listed below, including
 Laion2B-multi\footnote{\url{https://huggingface.co/datasets/laion/laion2B-multi-joined-translated-to-en}},
Chinese CLIP\footnote{\url{https://github.com/OFA-Sys/Chinese-CLIP}}, Russian CLIP\footnote{\url{https://github.com/ai-forever/ru-clip}}, Japanese CLIP\footnote{\url{https://github.com/rinnakk/japanese-clip}}, Korean CLIP\footnote{\url{https://github.com/jaketae/koclip}}, and Italian CLIP\footnote{\url{https://github.com/clip-italian/clip-italian}},
ProtoVision XL\footnote{\url{https://civitai.com/models/125703/protovision-xl-high-fidelity-3d-photorealism-anime-hyperrealism-no-refiner-needed}},
 OpenCLIP ViT-bigG\footnote{\url{https://huggingface.co/laion/CLIP-ViT-bigG-14-laion2B-39B-b160k}},
 SD1.5\footnote{\url{https://huggingface.co/runwayml/stable-diffusion-v1-5}} ,
 AltDiffusion-m9\footnote{\url{https://huggingface.co/BAAI/AltDiffusion-m9}}, 
 AltDiffusion-18\footnote{\url{https://huggingface.co/BAAI/AltDiffusion-m18}},
 GlueGen code\footnote{\url{https://github.com/salesforce/GlueGen}},
 ``xxmix9realistic''\footnote{\url{https://civitai.com/models/124421/xxmix9realisticsdxl}},\\
 ``samaritan-3d-cartoon''\footnote{\url{https://civitai.com/models/81270/samaritan-3d-cartoon}},
 ``csal\_scenery''\footnote{\url{https://civitai.com/models/118559/ancient-chinese-scenery-background-xl}},
 ``controlnet-canny''\footnote{\url{https://huggingface.co/diffusers/controlnet-canny-sdxl-1.0}},
SD-XL Inpainting 0.1\footnote{\url{https://huggingface.co/diffusers/stable-diffusion-xl-1.0-inpainting-0.1}},
Latent Consistency Models\cite{luo2023lcmlora}\footnote{\url{https://huggingface.co/latent-consistency/lcm-lora-sdxl}},
Segmind Stable Diffusion Model\footnote{\url{https://huggingface.co/segmind/SSD-1B}}.
SDXL-Turbo Model\footnote{\url{https://huggingface.co/stabilityai/sdxl-turbo}}.
ChatGLM-6B \footnote{\url{https://huggingface.co/THUDM/chatglm3-6b}},
Qwen-7B \footnote{\url{https://huggingface.co/Qwen/Qwen-7B}},
Baichuan2-7B \footnote{\url{https://huggingface.co/baichuan-inc/Baichuan2-7B-Base}},
Palyground V2 \footnote{\url{https://huggingface.co/playgroundai/playground-v2-1024px-aesthetic}},
Kandinsky V3 \footnote{\url{https://huggingface.co/kandinsky-community/kandinsky-3}},
Stable Cascade \footnote{\url{https://huggingface.co/stabilityai/stable-cascade}},
In addition, we also plan to release our code.

\section{Detailed Multilingual Performance}
\label{Multilingual}

\subsection{Performance with Chinese}
\label{MG_ZH}
\begin{figure}[ht]
\centering
\subfloat[Training data with resolutions around 640.]{
		\includegraphics[scale=0.38]{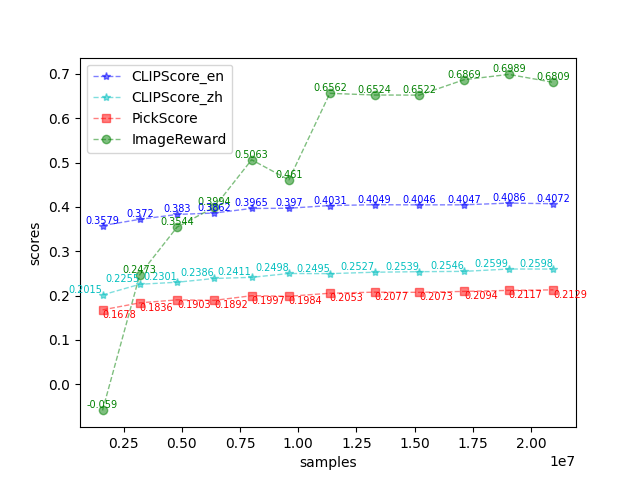}\label{fig_eval_zh_a}}
\subfloat[Training data with resolutions around 1024.]{
		\includegraphics[scale=0.38]{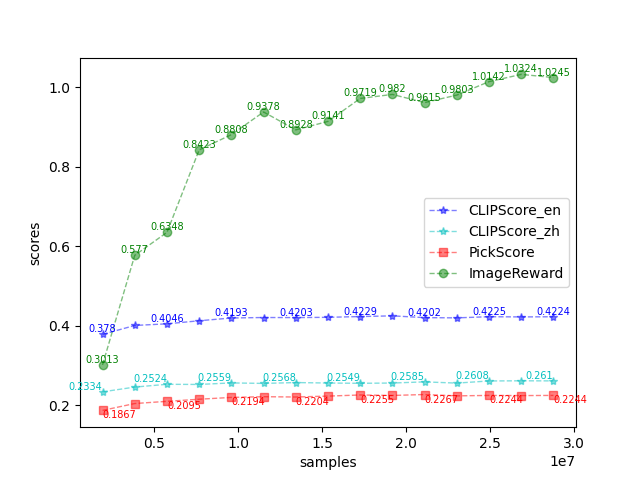}\label{fig_eval_zh_b}}
\caption{The evaluation results of training data with different resolutions in Chinese language.}
\label{fig_eval_zh}
\end{figure}

The evaluation results of training data with different resolutions in Chinese language is shown in Figure~\ref{fig_eval_zh}, where ``CLIPScore\_en'' represents the evaluation results based on the OpenCLIP ViT-bigG model for MG data, Two evaluation indicators based on human feedback, PickScore and ImageReward, are also evaluated on MG data. ``CLIPScore\_zh'' represents the evaluation results based on the Chinese CLIP model for MC data.
Since the final stage training data resolution for SDXL is bucketed around 1024, Figure~\ref{fig_eval_zh_b} shows that for the results based on Chinese data training with 1024 resolution, the ImageReward metric can consistently remain at around 1.

In order to explore the impact of training data with different resolutions on the final results, we conduct bucketed training based on resolutions around 640. As shown in Figure~\ref{fig_eval_zh_a}, the SDXL Chinese T2I model trained on low-resolution data perform poorly on the general evaluation set MG, proving that the PEA-Diffusion strategy has certain requirements for the training data. The resolution of the data fed into the model should be consistent with that of the teacher model in order to maintain the performance of PEA-Diffusion in general evaluation.

\subsection{Quantitative Analysis}
\subsubsection{Quantitative Analysis with Other Language}

From Table~\ref{data_tab}, we can see that the other four language datasets have fewer data when bucketed around 1024 resolutions. Therefore, we used the data around 640 resolutions for training.

\begin{figure}[htbp]
	\centering
	\begin{minipage}{0.49\linewidth}
		\centering
		\includegraphics[width=0.9\linewidth]{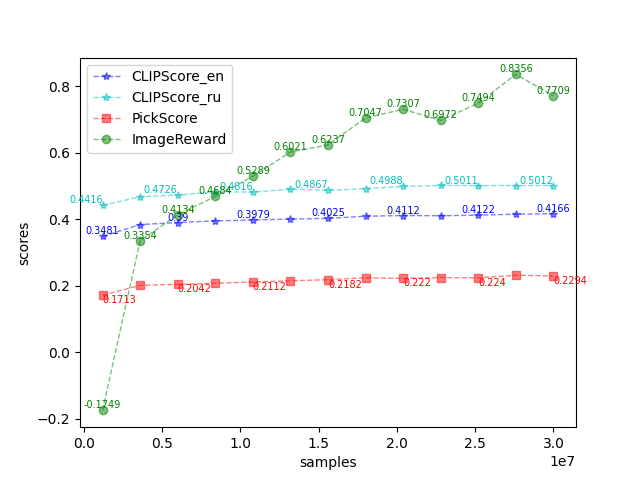}
		\caption{Russian evaluation results.}
		\label{ru}
	\end{minipage}
	\begin{minipage}{0.49\linewidth}
		\centering
		\includegraphics[width=0.9\linewidth]{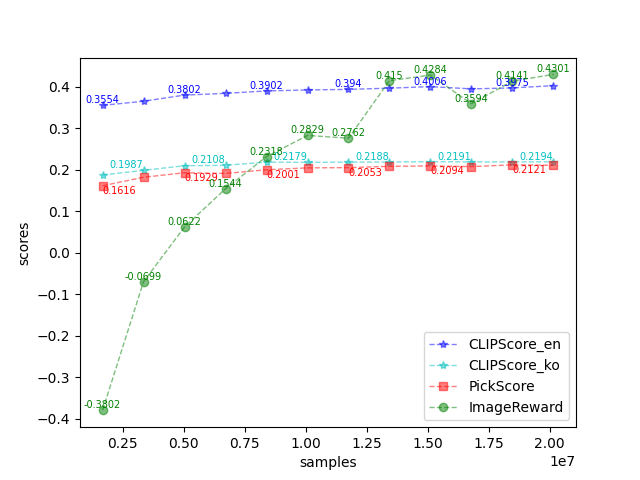}
		\caption{Korean evaluation results.}
		\label{ko}
	\end{minipage}
	
	\begin{minipage}{0.49\linewidth}
		\centering
		\includegraphics[width=0.9\linewidth]{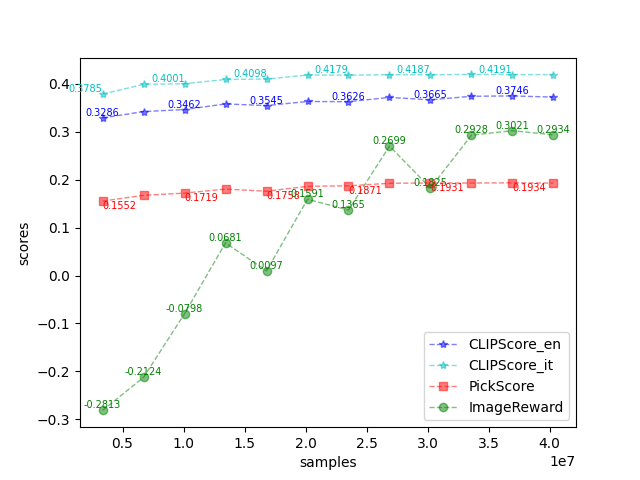}
		\caption{Italian evaluation results.}
		\label{it}
	\end{minipage}
	\begin{minipage}{0.49\linewidth}
		\centering
		\includegraphics[width=0.9\linewidth]{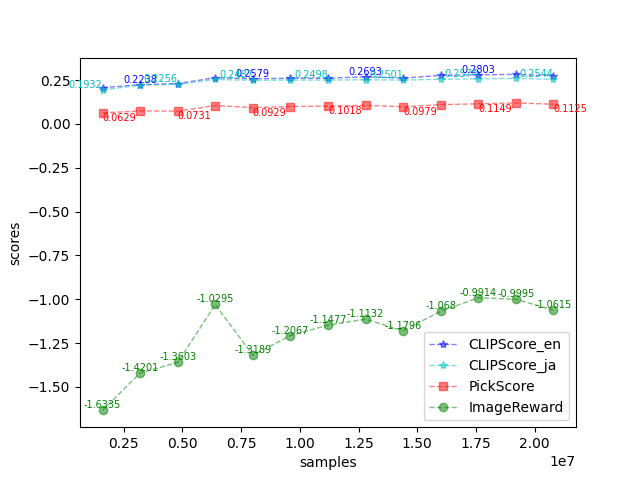}
		\caption{Japanese evaluation results.}
		\label{ja}
	\end{minipage}
\end{figure}

Table~\ref{data_tab} shows that Russian is the second largest language data in our collection. On the general evaluation dataset MG. Figure~\ref{ru} shows the ImageReward score for Russian data can remain stable at around 0.77, which is higher than the results of training Chinese data. Moreover, Sec. 4.3 Table 1 indicates that the CLIPScore on the MC evaluation dataset for Russian is higher than the results with the translation-based model by approximately 2.5 percentage points.

Table~\ref{data_tab} shows that the amount of Korean data used for training is only 1.6M. Figure~\ref{ko} shows the ImageReward score on the general evaluation dataset MG can still remain stable at around 0.43. Additionally, Sec. 4.3 Table 1 shows that on the MC evaluation dataset for Korean, the CLIPScore is higher than the results with the translation-based model by approximately 1.5 percentage points.

Table~\ref{data_tab} shows that the amount of Italian data used for training is only 18M. Figure~\ref{it} shows the ImageReward score on the general evaluation dataset MG can still remain stable at around 0.3. Additionally, Sec. 4.3 Table 1 shows that on the MC evaluation dataset for Italian, the CLIPScore is higher than the results with the translation-based model by approximately 5 percentage points.

The results for Japanese are not satisfactory as the training data amounted to 18M, but Figure~\ref{ja} shows the ImageReward and CLIPScore performance on both the general evaluation dataset MG and the MC evaluation dataset for Japanese is poor. We attribute this to the weak encoding capability of the CLIP model for Japanese that cannot be aligned with the SD model through the adapter layer, highlighting the high requirement of our proposed approach for the encoding capability of CLIP.

\subsubsection{Quantitative Analysis with Specified CLIPScore}
As shown in Table~\ref{tab:CLIP_five}, we also calculate the CLIP scores of the five languages using the CLIP model with corresponding languages in the MG teseset. It can be observed that out PEA-Diffusion still outperforms translation methods in terms of Chinese, Italian and Japanese while is only slightly inferior on Russian and Korean. This phenomenon demonstrates that translation method will lose critical information even for general prompts, leading to image-text matching degradation. Our PEA-Diffusion avoids the translation procedure while generates images based on the original prompts by an end-to-end manner, which will be a more optimal strategy. 
\begin{table}[!ht]
\centering
 \caption{\small{CLIPScore of five languages in MG data.}}
\resizebox{1\linewidth}{!}{
\begin{tabular}{llllll}
\hline
Methods       & Chinese   & Russian  & Italian   & Korean & Japanese \\ \hline
Translate     & 0.3206          & 0.5603        & 0.5166          & 0.2946        & 0.1819        \\
PEA-Diffusion & \textbf{0.3242} & 0.5321        & \textbf{0.5398} & 0.2870        & \textbf{0.2189}      \\ \hline
\label{tab:CLIP_five}
\end{tabular}}
\end{table}

\subsection{Qualitative Analysis}
\label{Other_Language}
\begin{figure*}[!h]
	\centering
    \setlength{\abovecaptionskip}{1em}
	\includegraphics[width=1\textwidth]{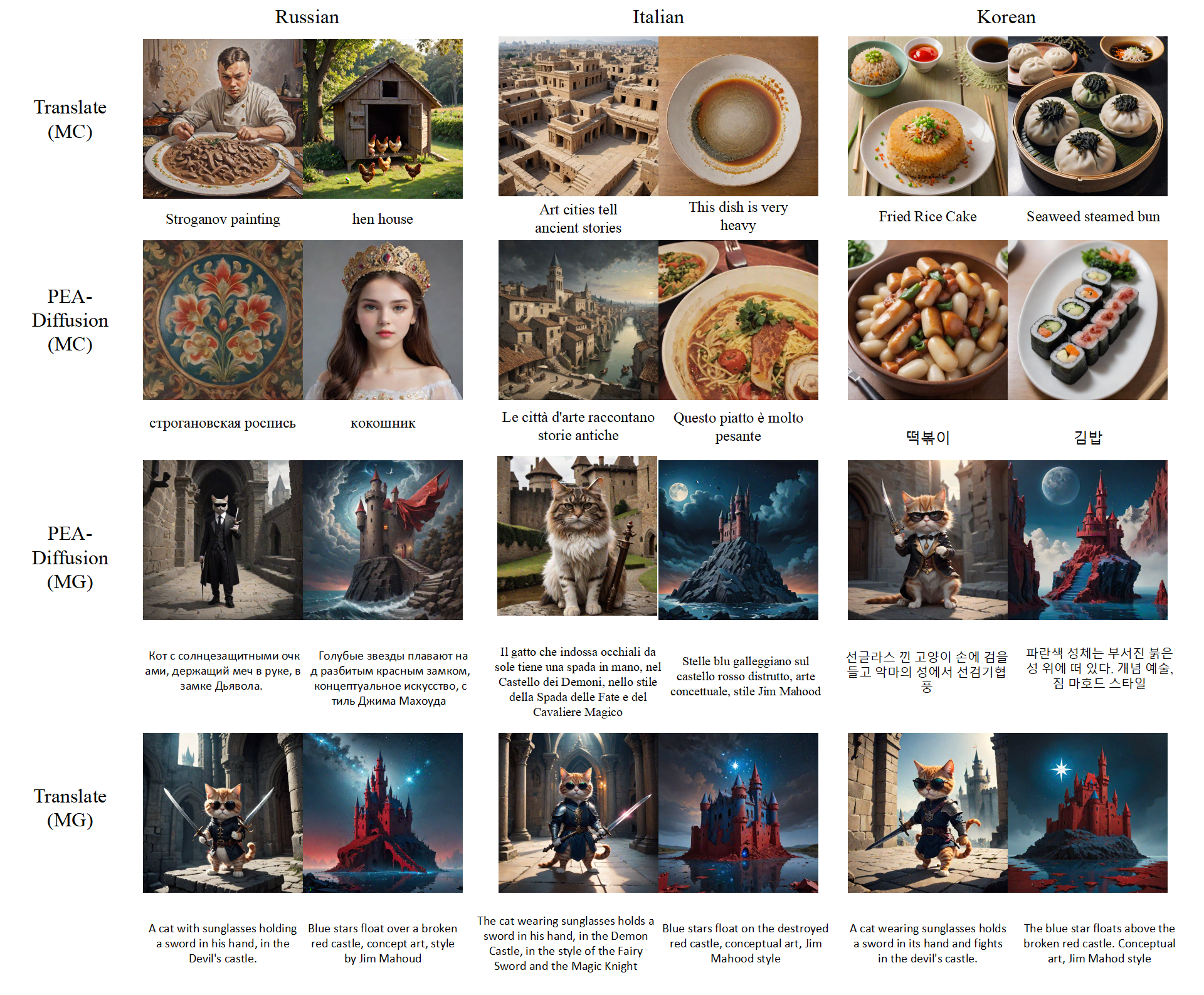}
 \caption{\small{Qualitative analysis with different languages.}}
\label{fig_other_la}
\end{figure*}

Figure~\ref{fig_other_la} provides a comparison of generated images for Russian, Italian, and Korean when compared to translations for MC and MG data prompts. The main issue with MC data is still related to the gap caused by the ambiguity in translations. Meanwhile, on MG data, we observed a significant loss of semantic information in Italian language prompts, which is consistent with the objective metrics shown in Figure~\ref{it}. However, the performance for Russian and Korean was relatively better. Due to the weaker representational ability of CLIP for Japanese language, images were not presented for this language.

\section{More Ablation Experiments}
\label{Hybrid}

\subsection{Hybrid Training Ablation}
For Chinese, we additional conduct two experiments: non-hybrid training with Multilingual-CLIP and non-hybrid training with Chinese-CLIP.

\textbf{Non-hybrid Training with Multilingual-CLIP.}
Non-hybrid training with Multilingual-CLIP is a natural solution because it is difficult for multilingual CLIP itself to encode cultural concepts unique to each language, so there is no need for hybrid training. In addition, only single-language English data is needed for training, which has great advantages in terms of training cost, data acquisition, and unified multilingual T2I.

From Table~\ref{Hybrid_tab}, we can see that for Chinese, the ``CLIPScore\_zh'' metric on the MC evaluation dataset is slightly higher than the translated CLIPScore\_zh score. More importantly, the performance on the general evaluation dataset MG is almost just about right with the translated score. Although there is a difference of 0.15 in the ImageReward metric, we have actually observed that as the ImageReward score improves, the significance represented by the difference in scores above 1 becomes smaller.

\textbf{Non-hybrid Training with Chinese-CLIP.}
For the non-hybrid training with Chinese-CLIP, we use only the Chinese-English parallel data, and to ensure data quality, the English captions are generated using BLIP2\cite{li2023blip2} while the corresponding Chinese descriptions are translated from English using a translation tool. This is because BLIP2 itself struggles with generating English descriptions with Chinese context, and adding the translation process guarantees that the translated Chinese descriptions do not contain any Chinese context. Moreover, we sample 500 text descriptions and find none related to the Chinese context.

From Table~\ref{Hybrid_tab}, it can be seen that for Chinese data without hybrid training, the ``CLIPScore\_zh'' is still almost one point higher compared to the results from the translation, and 0.5 points higher compared to the results from the multilingual-CLIP model. This further indicates that the CLIP text encoder is a major factor affecting the output of cultural relevance understanding.
Since hybrid training adds Chinese language context image-text pairs to the training data and the training objective is $\mathcal L_{SD}$, the results are naturally the best. 

Of course, it is undeniable that in addition to the ability of CLIP itself, the hybrid training strategy has indeed improved the cultural-related image generation capabilities, to some extent sacrificing the general evaluation metrics. We will continue to work in this direction and strive to achieve the optimal evaluation metrics for both types of data.

\begin{figure}[!h]
\centering
\setlength{\belowcaptionskip}{0.5em}
\setlength{\abovecaptionskip}{0.5em}
\captionof{table}{\small{The impact of hybrid training with Chinese.}}
\label{Hybrid_tab}
\resizebox{1\linewidth}{!}{
\begin{tabular}{ccccc}
\toprule
\textbf{Evaluation} & \textbf{ImageReward} & \textbf{PickScore} & \textbf{CLIPScore\_en} & \textbf{CLIPScore\_zh} \\\midrule
Translate & 1.3647 & 0.2402 & 0.4401 & 0.2382  \\\midrule
\makecell{ w/ Multilingual-CLIP} & 1.2122 & 0.2324 & 0.4321 & 0.2407  \\
\makecell{PEA-Diffusion \\ w/o Hybrid Training} & 1.1020 & 0.2236 & 0.4258 & 0.2460  \\\midrule
PEA-Diffusion & 1.0245 & 0.2244 & 0.4224 & 0.2610  \\\bottomrule
\end{tabular}}
\end{figure}

\subsection{Details of Ablation Studies with 4.4 Section in Paper}
\begin{figure}[!h]
	\centering
    \setlength{\abovecaptionskip}{1em}
	\includegraphics[width=0.8\textwidth]{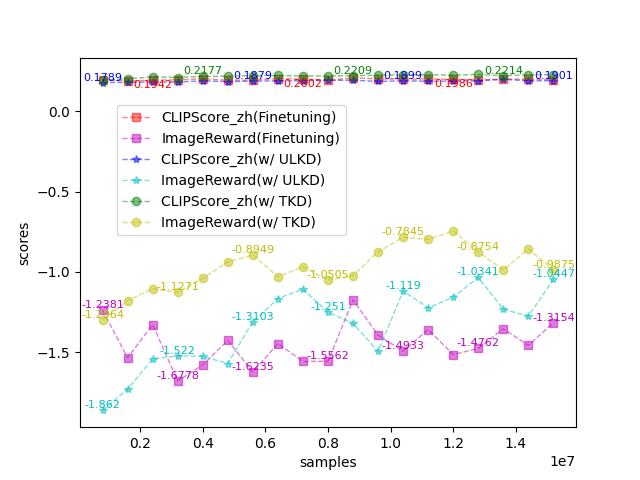}
 \caption{\small{Ablation results as training samples increase.}}
\label{fig_abla}
\end{figure}

Figure~\ref{fig_abla} shows the detailed results of the three ablation experiments on two metrics. Only fine-tuning without adding any KD results in the lowest overall performance, followed by adding UNet logit distillation (ULKD), which has a poor effect but is slightly better than not adding any knowledge distillation. Finally, we validated the KD in the text encoder (TKD). The overall performance of all three ablation experiments is very low, further proving that UNet feature distillation (UFKD) is the critical core of the entire PEA-Diffusion work.

As PickScore and ``CLIPScore\_en'' are relatively close, to better visualize the results, we only presented two metrics: the ``CLIPScore\_zh'' reflecting the MC dataset, and the ImageReward metric with human preferences reflecting the MG dataset.

\section{Low-cost Experiments}
\label{Convergence}
We analyze low-cost from two perspectives. The first is the relationship analysis between the PEA parameter size and other experimental parameters. The second is the comparison of the T2I method based on other low-cost fine-tuning methods.

\subsection{Analysis of PEA Parameter}
\begin{figure}[!h ]
	\centering
    \setlength{\abovecaptionskip}{1em}
	\includegraphics[width=0.8\textwidth]{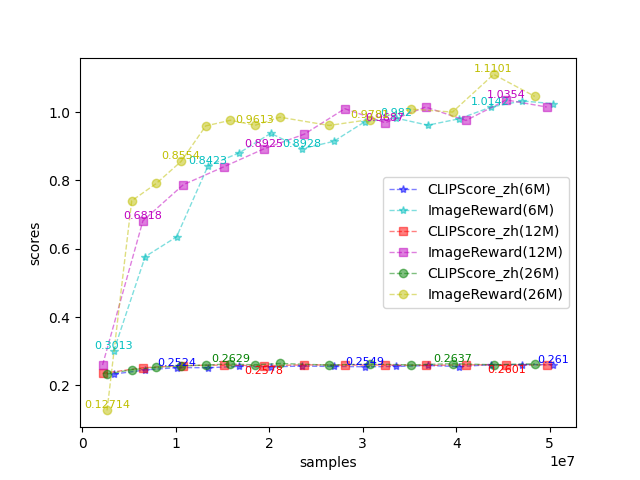}
 \caption{\small{Low-cost experiments with different PEA parameter.}}
\label{fig_cost}
\end{figure}

We focus on exploring the relationship between different training data volumes, the number of parameters in the adapter, and the evaluation metrics of MG and MC. We selected high-quality Chinese training data around 1024 resolution buckets, and then increased the number of parameters by modifying the layers and neuron numbers of the adapter to observe the changes in various metrics with an increase in the number of training samples. As shown in Figure~\ref{fig_cost}, a notable observation is that, for the ImageReward metric of the MG evaluation data, under the same training sample conditions, the ImageReward increases with an increase in the number of adapter parameters. This indicates a proportional relationship between the convergence speed of the model and the number of adapter parameters.

\subsection{Fine-tuning}
\label{Fine-tuning}
As a low-cost experiment for comparison, we also finetune the model to transfer language capabilities. Fine-tuning an image generation model can often lead to catastrophic forgetting, which significantly degrade the model's open-domain performance. To address these issues, we adopt a parameter-efficient fine-tuning strategy that preserves the original model's ability to the greatest extent possible.
During the learning stage, we only update the network parameters required for learning language character generation while freezing other parameters.
We adopt two fine-tuning schemes here. The first on is
Fine-tuning K and V matrices of the attention laters in the UNet.  As proposed in~\cite{kumari2023multi}, when updating the mapping from given text to image distribution, only updating $W_K, W_V$ in each cross-attention block $i$ is sufficient since text features are the only input to the key and value projection matrix.
The second one is fine-tuning with LoRA~\cite{hu2021lora}.

\begin{figure}[!h]
\centering
\setlength{\belowcaptionskip}{0.5em}
\setlength{\abovecaptionskip}{0.5em}
\captionof{table}{\small{Low-cost comparison of different strategies for Chinese language.}}
\label{Low_cost}
\resizebox{1\linewidth}{!}{
\begin{tabular}{cccc}
\toprule
\textbf{Evaluation} & \textbf{only $W_k^{(i)}$ and $W_v^{(i)}$} & \textbf{LoRA} & \textbf{PEA-Diffusion} \\\midrule
CLIPScore\_zh & 0.2485 & 0.2366 & 0.2610  \\
ImageReward & 0.0293 & -0.3214 &  1.0245 \\
CLIPScore\_en & 0.3854 & 0.3577 & 0.4224  \\
PickScore & 0.1758 & 0.1613 & 0.2244 \\\bottomrule
\end{tabular}}
\end{figure}

Table~\ref{Low_cost} shows that for the four metrics evaluated for MC and MG, the results of only fine-tuning the $W_K, W_V$ matrices are better than those obtained by LoRA training, but the overall scores are still relatively low. These results suggest that conventional fine-tuning methods may not effectively achieve language transfer in T2I synthesis.

\section{Performance of LLMs as Text Encoder for T2I}
\label{LLM_suppl}

\begin{figure}[tb!]
\centering
\captionof{table}{\small{Performance of LLMs as text encoder for T2I in Chinese specific language.}}
\label{LLM}
\resizebox{1\linewidth}{!}{
\begin{tabular}{cccccc}
\hline
Model Base            & X-encoder    & ImageReward & PickScore & CLIPScore\_en & CLIPScore\_zh \\ \hline
\multirow{3}{*}{SDXL} & ChatGLM-6B  & 0.8547      & 0.2105    & 0.3985        & 0.2454        \\
                      & Qwen-7B   & 0.7547      & 0.2005    & 0.3918        & 0.2398        \\
                      & Baichuan2-7B  & 0.7952      & 0.2014    & 0.3907        & 0.2354        \\ \hline
Palyground V2     & ChatGLM-6B   & 0.8924      & 0.2154    & 0.4014        & 0.2415        \\ \hline
Kandinsky V3       & ChatGLM-6B   & 0.6751      & 0.1928    & 0.4032        & 0.2401        \\ \hline
Stable Cascade   & ChatGLM-6B   & 0.6450      & 0.2014    & 0.3987        & 0.2584        \\ \hline
\end{tabular}}
\end{figure}

To further enhance the performance of PEA-Diffusion on non-English text, we attempted to replace the encoder with a stronger X-Test encoder. Large Language Models (LLMs) have demonstrated exceptional semantic understanding and generation capabilities. Although LLMs are mostly designed as decoder-only structures, our experiments on Chinese indicate that LLMs can also serve as text encoders for SD through PEA adaptation. Table \ref{LLM} presents the experimental results. We experimented with three different base models and four large language models.

\begin{figure*}[htbp]
	\centering
    \setlength{\abovecaptionskip}{1em}
	\includegraphics[width=0.95\textwidth]{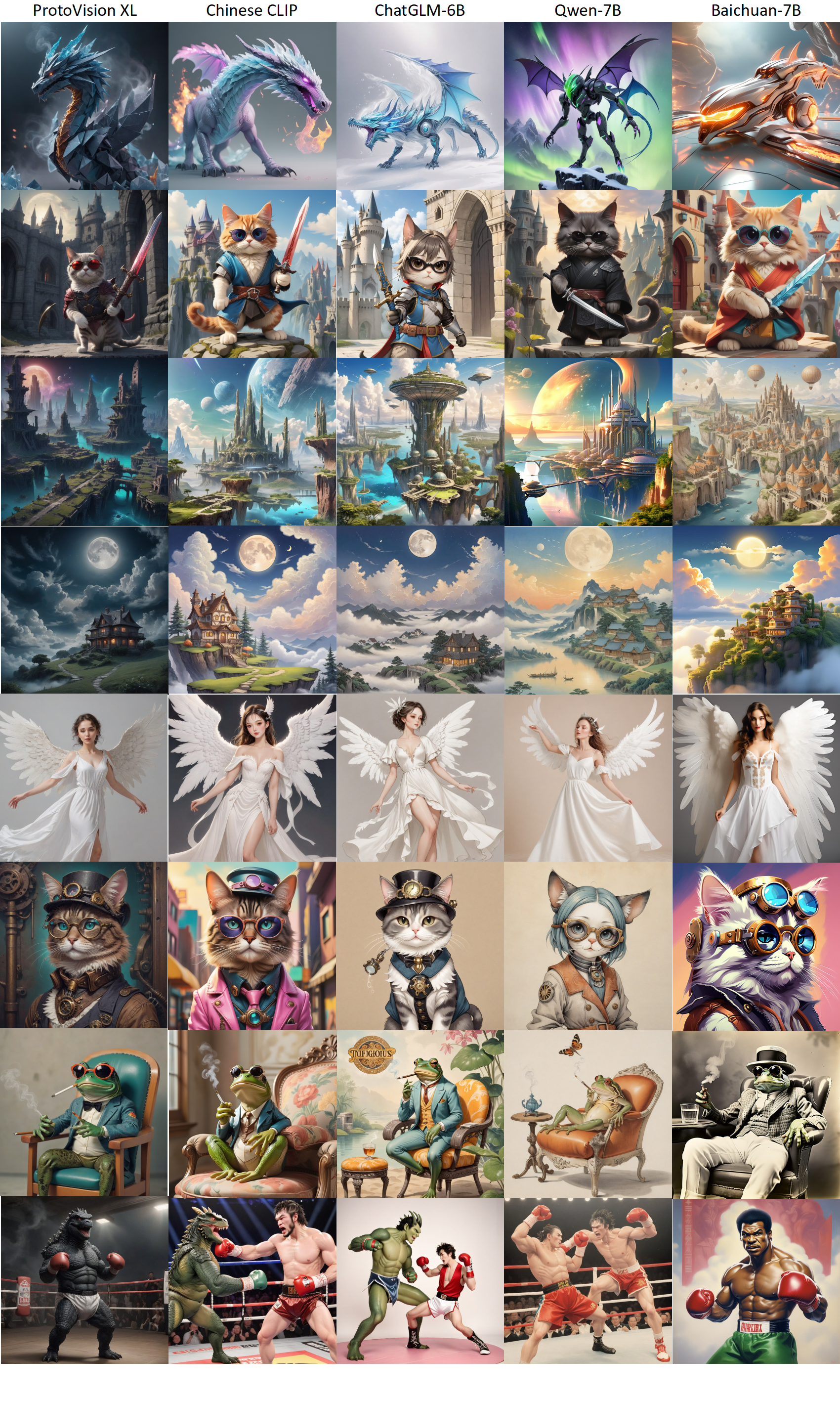}
 \caption{\small{More qualitative results with LLMs as text encoder for T2I.}}
\label{llm_result}
\end{figure*}

Figure~\ref{llm_result} showcases the effects of different models used as T2I text encoders, with the emphasize that the exhibited results are randomly selected. 
The first column is teacher model ProtoVision XL that based on SDXL and undergoes high-quality fine-tuning. The second column corresponds to PEA diffusion trained based on the Chinese CLIP. The subsequent three columns depict PEA diffusion trained with different LLMs. The eight prompts tested here are:
\begin{itemize}
\item  3D Technological, Futuristic, Cyberpunk, Aurora, Smoke, Ice Dragon, Origami.
\item A cat with sunglasses holds a sword in his hand, in a demon castle, in the style of Sword and Fairy.
\item Grandiose cosmic backdrops, city ruins on floating islands, fantasy concept art, futurism.
\item  Landscape, Clouds swirled around the house, Moon, World, Fantasy, mythology.
\item A winged woman in a white dress.
\item A steampunk cat with cool glasses, Portait, is all the rage in the art world, vaporwave art.
\item  A very detailed, realistic photograph of a frog wearing small sunglasses and smoking a cigarette in a beautiful mid-century modern style chair, designed by Sonia Delaunay. Beautiful colors. Vintage film camera.
\item Godzilla became a professional boxer.
\end{itemize}
The four columns is the specific language input corresponding to X-Test encoder. Although there exists some disparity between the Imagereward in Table~\ref{LLM} and the teacher model, the qualitative subjective gap in visual effects is relatively small when viewing the overall presentation in Figure~\ref{llm_result}. The only exception is a few instances where the semantic alignment effects are slightly inferior, such as the last row of images in Figure~\ref{llm_result}.

\begin{figure*}[hbp]
	\centering
    \setlength{\abovecaptionskip}{1em}
	\includegraphics[width=1\textwidth]{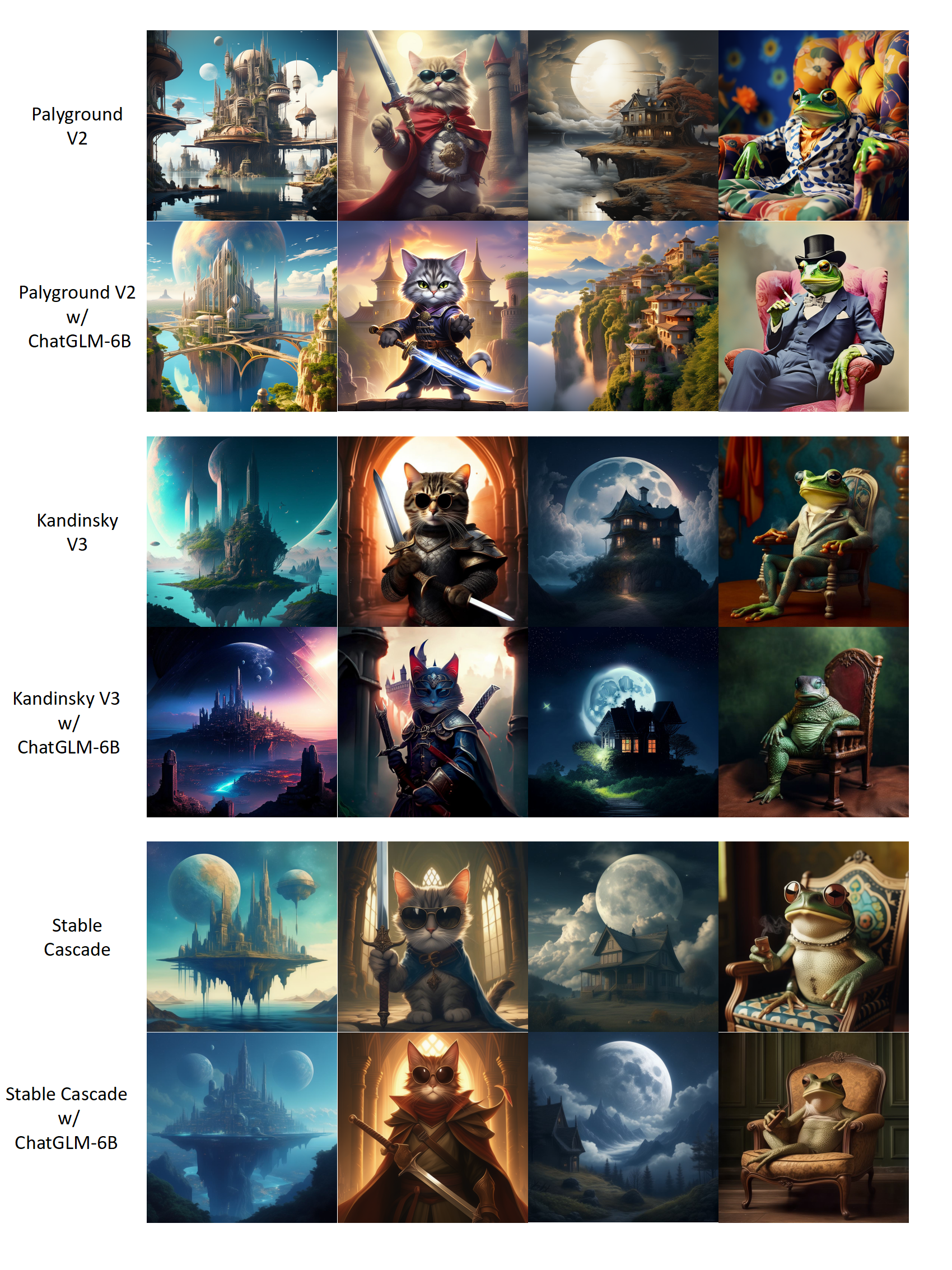}
 \caption{\small{Qualitative results of different model base. The implementation based on three different framework models demonstrates the flexible adaptability of PEA diffusion.}}
\label{llm_result1}
\end{figure*}

Figure~\ref{llm_result1} illustrates the performance of different backbone frameworks when using ChatGLM-6B as the T2I encoder. The conclusions are similar to Figure~\ref{llm_result} and further demonstrate the plug-and-play capabilities of the PEA method, allowing for rapid adaptation to different model frameworks.

\section{More Qualitative Results for Sec. 4.3 Table 1}
\label{table1}

Although the CLIPScore for MC in Sec. 4.3 Table 1 shows some improvement in specific language, the comparison of MG data against the performance of the teacher model is not particularly ideal. The ImageReward and PickScore metrics both show a slight decrease. Therefore, it is necessary to analyze the qualitative results.

\begin{figure*}[htbp]
	\centering
    \setlength{\abovecaptionskip}{1em}
	\includegraphics[width=1\textwidth]{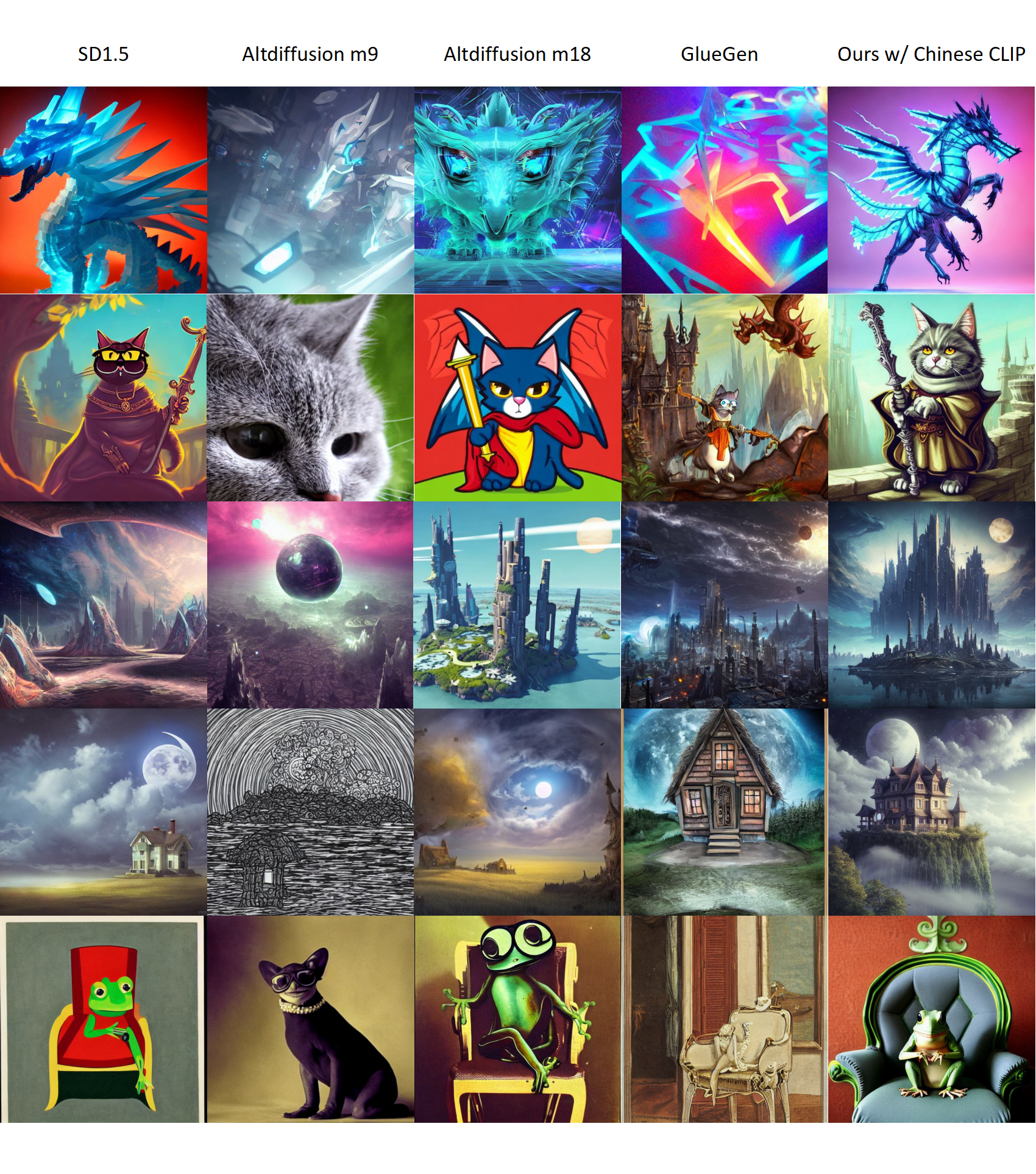}
 \caption{\small{Qualitative result comparisons in Sec. 4.3 Table 1 with SD1.5 model base.}}
\label{table_sd}
\end{figure*}

Firstly, for the model base results in Figure~\ref{table_sd} targeting SD1.5, it is evident that our method has a significant advantage over Altdiffusion and GlueGen. Compared to the teacher model, there is a decrease of 0.1442 in the ImageReward metric, indicating a slight decline in image quality. However, this situation improves significantly when SDXL is used as the model base.

\begin{figure*}[htbp]
	\centering
    \setlength{\abovecaptionskip}{1em}
	\includegraphics[width=1\textwidth]{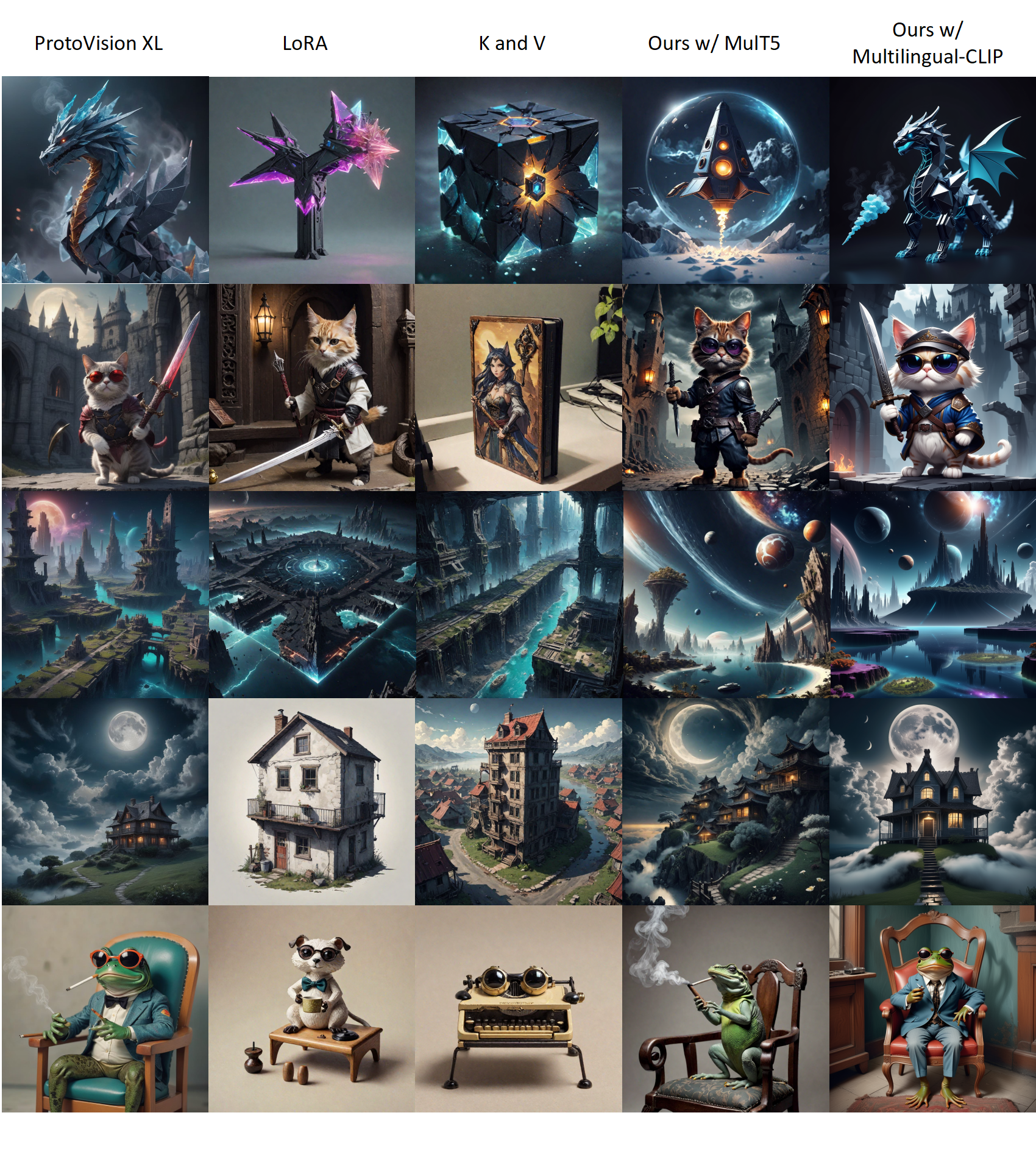}
 \caption{\small{Qualitative result comparisons in Sec. 4.3 Table 1 with SDXL model base.}}
\label{table_sdxl}
\end{figure*}

In Figure~\ref{table_sdxl}, it can be observed that both LORA and $W_K, W_V$ fine-tuning show a noticeable decrease in image quality. However, the last two columns, based on PEA strategies, especially the last column employing Multilingual-CLIP as PEA diffusion, exhibit a difficulty in qualitatively discerning the decline in image quality, despite the ImageReward metric showing a decrease of 0.1525. The same reasoning applies to the first two columns in Figure~\ref{llm_result}.

Here, we speculate that beyond a certain threshold of the ImageReward metric, possibly around 1 or greater, the differences in the metric may not necessarily translate into significant differences in actual outcomes. Of course, we will further validate this speculation through additional experiments.

\section{User Study}
\label{user_study}

\begin{table}[ht]
\captionof{table}{\small{User studies.}}
\label{studies}
\resizebox{1\linewidth}{!}{
\begin{tabular}{lllllll}
\hline
                          & \multicolumn{2}{l}{visual quality}                                              & \multicolumn{2}{l}{semantic alignment}                                          & \multicolumn{2}{l}{success rate}                                                \\ \cline{2-7} 
\multirow{-2}{*}{Methods} & MC                                     & MG                                     & MC                                     & MG                                     & MC                                     & MG                                     \\ \hline
Translate                 & {2.2267}          & {3.6126}          & {1.2981}          & {\textbf{3.2042}} & {0.2577}          & {\textbf{0.8178}} \\ \hline
LoRA                      & {0.6536}          & {1.6889}          & {0.8727}          & {0.9202}          & {0.1639}          & {0.1590}          \\ \hline
PEA-Diffusion             & {\textbf{2.8423}} & {\textbf{3.6186}} & {\textbf{2.3539}} & {3.0643}          & {\textbf{0.4978}} & {0.8089}          \\ \hline
\end{tabular}}
\end{table}

We generated 2000 images, based on a total of 400 prompts from MG and MC, with each prompt producing 5 images to determine the success rate. We then established specific scoring criteria and assigned the human evaluation to over ten experienced annotators. Their results were averaged to form the final result. The first two indicators were graded on a scale ranging from 0 to 4, with five levels in total. The higher the score, the better the performance. As shown in Table~\ref{studies}, our method exhibits a significant advantage over translation results and LoRA in three metrics for the MC dataset, particularly in semantic alignment and success rate. For the MG dataset, it is worth highlighting that there is no disparity between our method and translation regarding visual quality, while both semantic alignment and success rate are only marginally lower. This further substantiates that our approach's efficacy does not diminish on general data. More details can be found in Appendix ~\ref{table1}.

\section{Downstream Experiment}
\label{Adaptability}
\subsection{LCM Quantitative Analysis}
Latent Consistency Models~\cite{luo2023latent} have achieved impressive performance in accelerating T2I generative tasks, producing high-quality images with minimal inference steps. ``lcm-lora-sdx'' is a universal acceleration module based on SDXL that allows to reduce the number of inference steps to only between 2 and 8 steps.

\begin{figure*}[htbp]
	\centering
	\includegraphics[width=1\textwidth]{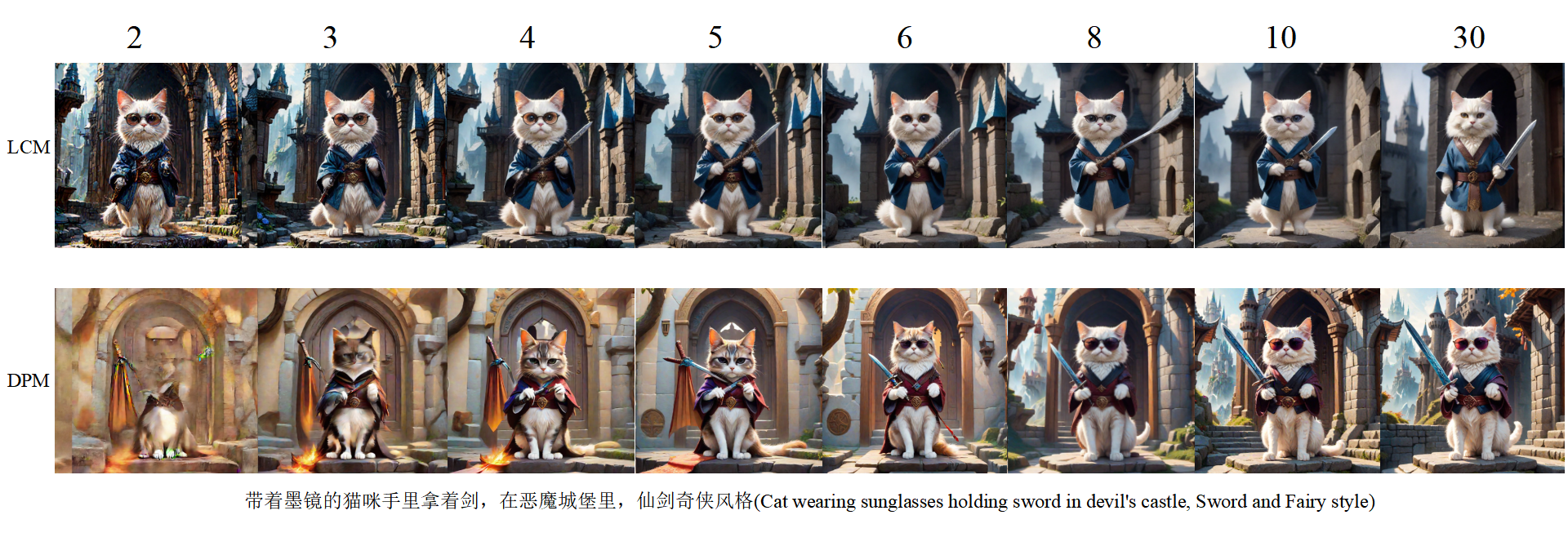}
 \caption{\small{Qualitative analysis of PEA-Diffusion in LCM for training Chinese language in SDXL based model.}}
\label{fig_lcm}
\end{figure*}

As an example of LCM, we evaluate the PEA-Diffusion model for Chinese training with the SDXL-based model, using quantitative metrics to measure its adaptability. Table~\ref{tab:chinese_lcm} shows the results for different sampling steps, ranging from 2 to 8 steps and 10, 20, and 30 steps as evaluated by MG, as well as the results based on a single MG evaluation metric. Our previous experiments used a DPM\cite{song2021scorebased} sampler with a step of 30. The important result we found through experimental comparison is that LCM-LoRA achieved optimal performance with just 3 sampling steps. Furthermore, compared with the Chinese PEA-Diffusion model, the DPM sampler only decreases by 0.3 points with 30 steps, which can be observed in Figure~\ref{fig_lcm} and show no significant difference in subjective results. CLIPScore\_en decreases by 0.15 points. However, the gap between these two metrics is likely due to the LCM strategy itself. Based on Table~\ref{tab:chinese_lcm} and Figure~\ref{fig_lcm}, we can conclude that the PEA module combined with LCM can be easily plugged in and applied.

\begin{figure*}[!ht]
\centering
\captionof{table}{\small{Quantitative indicators of PEA-Diffusion in LCM for training Chinese language in SDXL based model.}}
\label{tab:chinese_lcm}
\resizebox{1\linewidth}{!}{
\centering

\begin{tabular}{lllllllll}
\hline
\multicolumn{1}{c}{\multirow{2}{*}{steps}} & \multicolumn{4}{c}{LCM}                                                                                                            & \multicolumn{4}{c}{DPM}                                                                                                                 \\ \cline{2-9} 
\multicolumn{1}{c}{}                       & \multicolumn{1}{c}{ImageReward} & \multicolumn{1}{c}{PickScore} & \multicolumn{1}{c}{CLIPScore\_en} & \multicolumn{1}{c}{CLIPScore\_zh} & \multicolumn{1}{c}{ImageReward} & \multicolumn{1}{c}{PickScore} & \multicolumn{1}{c}{CLIPScore\_en} & \multicolumn{1}{c}{CLIPScore\_zh} \\ \hline
2                                          & 0.7689                          & 0.2151                        & 0.4094                            & 0.2545                            & -0.9963                         & 0.1066                        & 0.3725                            & 0.2325                            \\
\textbf{3}                                 & \textbf{0.8774}                 & \textbf{0.2207}               & \textbf{0.4077}                   & \textbf{0.2584}                   & 0.0515                          & 0.1558                        & 0.4024                            & 0.2525                            \\
4                                          & 0.8757                          & 0.2210                        & 0.4031                            & 0.2541                            & 0.4120                          & 0.1774                        & 0.4127                            & 0.2532                            \\
5                                          & 0.8593                          & 0.2184                        & 0.4019                            & 0.2586                            & 0.6140                          & 0.1939                        & 0.4229                            & 0.2601                            \\
6                                          & 0.8466                          & 0.2180                        & 0.3994                            & 0.2580                            & 0.7322                          & 0.2016                        & 0.4204                            & 0.2599                            \\
7                                          & 0.8602                          & 0.2173                        & 0.4021                            & 0.2586                            & 0.8657                          & 0.2070                        & 0.4214                            & 0.2598                            \\
8                                          & 0.9143                          & 0.2180                        & 0.4016                            & 0.2588                            & 0.8773                          & 0.2137                        & 0.4188                            & 0.2588                            \\
10                                         & 0.8428                          & 0.2157                        & 0.3987                            & 0.2559                            & 1.0036                          & 0.2244                        & 0.4124                            & 0.2575                            \\
20                                         & 0.7699                          & 0.2136                        & 0.3968                            & 0.2556                            & 1.0336                          & 0.2244                        & 0.4125                            & 0.2574                            \\
\textbf{30}                                & 0.7574                          & 0.2106                        & 0.3987                            & 0.2546                            & \textbf{1.0245}                 & \textbf{0.2244}               & \textbf{0.4224}                   & \textbf{0.2610}                   \\ \hline
\end{tabular}}
\end{figure*}

\subsection{More Downstream Generated Images}
\begin{figure*}[htbp]
	\centering
    \setlength{\abovecaptionskip}{1em}
	\includegraphics[width=1\textwidth]{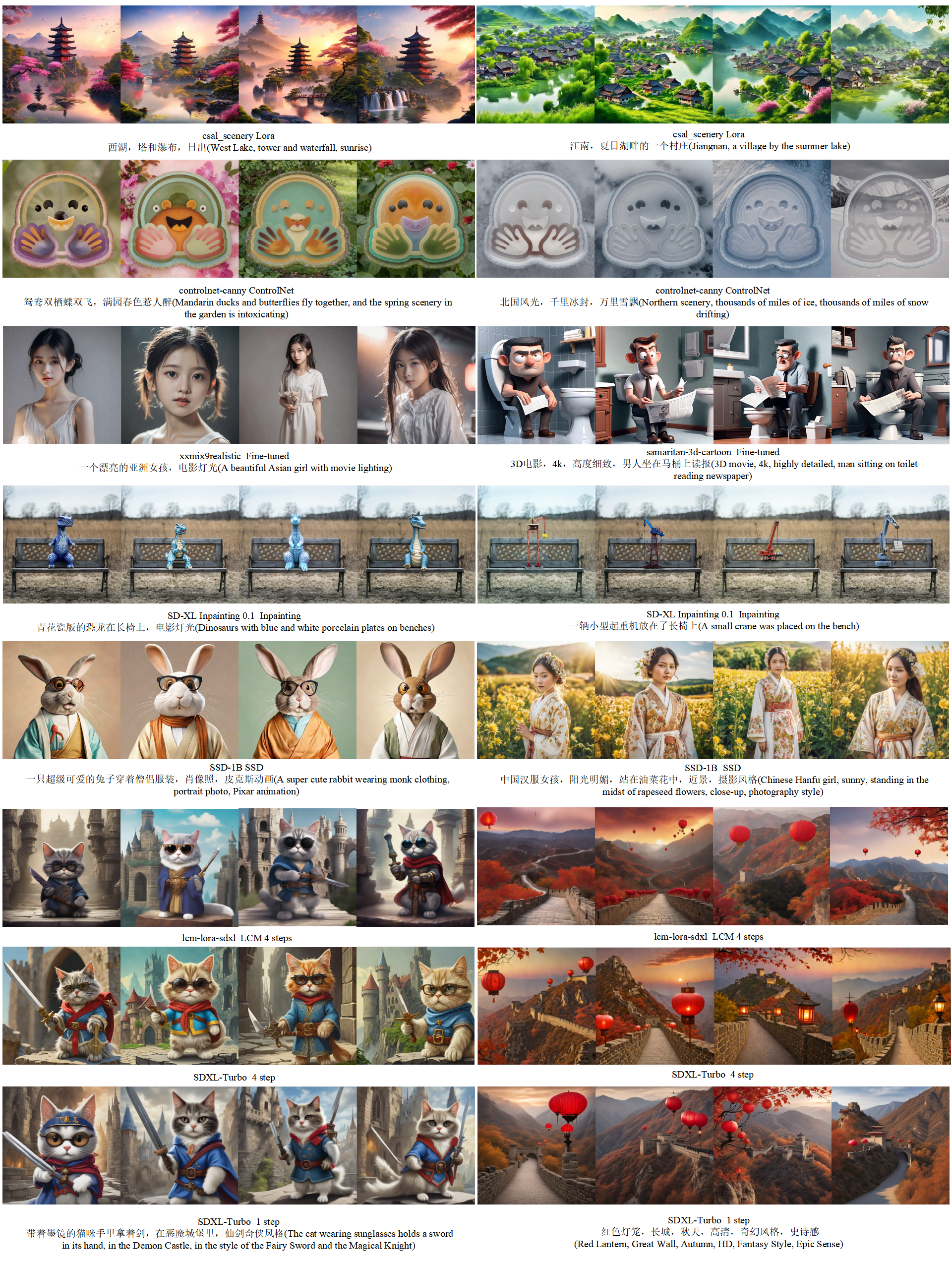}
 \caption{\small{More adaptability experiment.}}
\label{fig_downstream}
\end{figure*}

Figure~\ref{fig_downstream} shows more image generation situations of downstream adaptation such as fine-tuned checkpoint, LoRA, ControlNet, Inpainting, SSD, LCM, SDXL-Turbo. All the pretrained checkpoints will be detailed in Appendix~\ref{data}.
Through Figure~\ref{fig_downstream}, we have proven that the PEA module can be used as a plugin between the Chinese CLIP text encoder and the UNet, achieving language transfer ability from English to Chinese without any other process.

\clearpage  



\end{document}


\title{Supplementary Material for Fine-tuning Diffusion Models
    for Enhancing Face Quality in Text-to-image Generation} 


\author{First Author\inst{1}\orcidlink{0000-1111-2222-3333} \and
Second Author\inst{2,3}\orcidlink{1111-2222-3333-4444} \and
Third Author\inst{3}\orcidlink{2222--3333-4444-5555}}

\authorrunning{F.~Author et al.}

\institute{Princeton University, Princeton NJ 08544, USA \and
Springer Heidelberg, Tiergartenstr.~17, 69121 Heidelberg, Germany
\email{lncs@springer.com}\\
\url{http://www.springer.com/gp/computer-science/lncs} \and
ABC Institute, Rupert-Karls-University Heidelberg, Heidelberg, Germany\\
\email{\{abc,lncs\}@uni-heidelberg.de}}

\maketitle

\appendix
\section{Algorithm}
We leverage the face pairs for fine-tuning DMs. 
A corresponding mask $m$ indicates the location of the face in the image while the prompt $p$ provides a description of the content of the image.
The fine-tuning algorithm is presented in the~\cref{algo:ft_dm}.
We provide more results in~\cref{fig:more results}.
\begin{algorithm}[htbp]
    \caption{Fine-tune DMs by Face Pairs}
    \label{algo:ft_dm}
    \begin{algorithmic}[1]
        \STATE \textbf{Input}: Dataset $D = \{(z^g,z^b,m,p),\ldots\}$, the original DM $\epsilon_{\theta^*}(z_t,t)$, \\the target DM $\epsilon_{\theta}(z_t,t)$, the distance measure function $d$ 
        \STATE \textbf{Initialization}: The noise scheduler, the number of noise scheduler timestep $T$
        \FOR{each $(z^g,z^b,m,p)$ in $D$}
            \STATE $t \xleftarrow{} rand(0,0.5T), \epsilon \sim \mathcal{N}(0,I)$
            \STATE $z^g_t,z^b_t \xleftarrow{} Forward(z^g,t,\epsilon), Forward(z^b,t,\epsilon)$ // Add noise to images
            \STATE \textbf{no grad:} $\Bar{z}^g_0 \xleftarrow{} OneStepDenoise(z^g_t,\epsilon_{\theta^*}(z^g_t,t))$ // Predict $\Bar{z}^g_0$
            \STATE \textbf{with grad:} $\Bar{z}^b_0 \xleftarrow{} OneStepDenoise(z^b_t,\epsilon_{\theta}(z^b_t,t))$ // Predict $\Bar{z}^b_0$
            \STATE $L_{guidance} \xleftarrow{} d(m\circ\Bar{z}^g_0,m\circ\Bar{z}^b_0)$
            \STATE Update $\epsilon_{\theta}$ based on $L_{guidance}$
        \ENDFOR
        \RETURN The fine-tuned DM $\epsilon_{\theta}$
    \end{algorithmic}
\end{algorithm}

\section{More Ablation Studies on Timesteps}
In the paper, we conduct ablation studies on timesteps.
Here we provide more detailed ablation studies on them.
To explore the optimal strategy for achieving better face generation or detail generation, we consider different timestep splitting points $T_{end}$ and take 250 as a unit.
Specifically, we choose from timestep 0 to $T_{end}$, i.e., taking values from $250$, $500$, $750$, and $1000$. 
The experiments take RV5.1 as the base model and DINO feature distance as the measure.
Except for the $T_{end}$, we keep other settings unchanged and take the same comparison settings.

As shown in~\cref{tab:ablation time}, as $T_{end}$ increases, the detail generation performance, which we focus on and is indicated by \textit{Face Score}, exhibits an initial increase followed by a continuous decrease, and $T_{end}=500$ is optimal, so we adopt it as the default setting in other experiments.
The reason why the detail generation is better when $T_{end}=750$ compared to when $T_{end}=250$ may be attributed to the fact that $T_{end}=750$ includes the interval of $T_{end}=500$.
As for the other metrics, their focus is primarily on the overall aesthetic quality, which is not within our scope of concern, rather than the specific face details.

\begin{table}[ht]
    \centering
    \caption{A more detailed ablation study on timesteps.}
    \setlength{\tabcolsep}{6mm}
    \begin{tabular}{c|c|c|c|c}
    \toprule[1.5pt]
    $T_{end}$ & CLIP-Score $\uparrow$ & IR $\uparrow$ & HPS $\uparrow$ & \textit{FS} $\uparrow$ \\
    \midrule[1.2pt]
    250 & 0.3251 & 1.80 & 0.3403 & 2.8309 \\
    500 & 0.3253 & 1.77 & 0.3350 & \textbf{2.8653} \\
    750 & 0.3249 & 1.78 & 0.3306 & 2.8395 \\
    1000 & 0.3251 & 1.62 & 0.3376 & 2.6980 \\
    \bottomrule[1.5pt]
    \end{tabular}
    \label{tab:ablation time}
\end{table}

\section{Datasets}
To provide a more intuitive understanding of the dataset, we present some details about the process of dataset formation and extract a subset of samples to help readers gain a preliminary understanding of the format and quality of the dataset, and better comprehend the methods and results presented in our paper.

\subsection{Ranking Criteria for Evaluation}
We establish the following annotation rules for human annotators:
\begin{itemize}
    \item[$\bullet$] Discard those triplets if there are no valid faces in any image;
    \item[$\bullet$] Focus solely on the faces and do not need to consider the alignment between the prompt and the image, the aesthetic aspect of the image itself, or any other irrelevant factors;
    \item[$\bullet$] Prioritize the rationality of the face before considering its aesthetic aspect.
    \item[$\bullet$] Select the most frontal and representative face for comparison purposes in multi-person scenes.
\end{itemize}
We also provide more image triplets in~\cref{fig:evaluation dataset} to see the correlation between face quality and human preference.

\subsection{Face Pairs Construction}
In the inpainting pipeline of DMs, the noise factor is a hyper-parameter, which should be chosen carefully.
We intend to degrade the face to resemble the quality of a bad face generated by DMs.
Experimentally, we find that an increased noise factor leads to more pronounced degradation, and different sizes of faces exhibit varying levels of resistance to the degradation effects.
To this end, we choose various noise factors based on the proportion of the face in the image as shown in~\cref{tab:noise factor}.
We go through multiple rounds of data filtering, including the removal of faces that are too small and blurry, as well as other invalid faces.
We present more examples in the implicit human preference dataset of faces in~\cref{fig: ds}.
\section{Face Generation Quality}
\subsection{Evaluation Discussion}
We observe a positive correlation between the quality of generated faces and the aesthetic quality of their corresponding images, which is influenced by the model's capabilities.
This correlation can be explained by the fact that stronger models exhibit a higher capacity for generating finer details and explains why IR and HPS can perform better than random guesses on predicting human preference on faces,
though they are designed and trained on the whole image and focus on the global features.
\begin{table}[t]
    \centering
    \setlength{\tabcolsep}{3.5mm}
    \caption{Different face area ratios correspond to various noise factors.}
    \begin{tabular}{rlcccc}
    \toprule
    \multicolumn{2}{c}{Face area ratio} & (0, 0.5\%] & (0.5\%, 1\%] & (1\%, 10\%] & (10\%, 100\%] \\
    \midrule
    \multicolumn{2}{c}{Noise factor} & 0.06 & 0.1 & 0.2 & 0.4 \\
    \bottomrule
    \end{tabular}
    \label{tab:noise factor}
\end{table}
\subsection{\textit{Face Score} Discussion}
We provide additional examples in~\cref{fig:more fs}.
It can be seen that when $\textit{Face Score}$ is higher than about 4, the generated faces are generally plausible.
Moreover, the higher the score, the more attractive the faces tend to be, albeit in different ways. 
Conversely, when the score falls below 4 and continues to decline, the faces are no longer satisfactory or realistic.
At first, the faces may appear slightly blurry and exhibit some distortion in certain details.
Eventually, significant blurriness causes the faces to become unrecognizable.
Based on the relationship between the \textit{Face Score} and the quality of face generation, we can take it as a reliable automated evaluation metric for assessing the quality of face generation.
\begin{figure}[ht]
  \centering
  \begin{tabular}{ccccc}
    \toprule  
    \includegraphics[width=2.2cm]{morervl2/An older woman riding a train while sitting under it's window..jpg} 
    & \includegraphics[width=2.2cm]{morervl2/Black and white photograph of a man sitting at a bench..jpg} 
    & \includegraphics[width=2.2cm]{morervl2/Female tennis player preparing to serve the ball.jpg}
    & \includegraphics[width=2.2cm]{morervl2/Girl looks back as she fixes flower on boy's suit.jpg}
    & \includegraphics[width=2.2cm]{morervl2/Girl trying out a game at a trade show..jpg} \\
    \includegraphics[width=2.2cm]{morervl2/Man in dress shirt and orange tie standing inside a building..jpg}
    & \includegraphics[width=2.2cm]{morervl2/Man in motorcycle leathers standing in front of a group of bikes.jpg}
    & \includegraphics[width=2.2cm]{morervl2/MAN ON A CONTRAPTION, SURROUNDED BY A BICYCLE AND ANOTHER PERSON.jpg}
    & \includegraphics[width=2.2cm]{morervl2/the boy has on black sneakers and his trying to grab the frisbee.jpg}
    & \includegraphics[width=2.2cm]{morervl2/The man is holding the child and lighting the candle..jpg} \\
    \includegraphics[width=2.2cm]{morervl2/The man stands on fence railing beside a table with food on it..jpg}
    & \includegraphics[width=2.2cm]{morervl2/The young girl wearing red and black rides atop the white pony..jpg}
    & \includegraphics[width=2.2cm]{morervl2/there is a male tennis player about to hit the ball.jpg}
    &\includegraphics[width=2.2cm]{morervl2/there is a woman laying in a bed using a lap top.jpg}
    & \includegraphics[width=2.2cm]{morervl2/Three men stand next to a car with its hood open..jpg} \\
    \includegraphics[width=2.2cm]{morervl2/Two beautiful young women baking a turkey in a pan...jpg} & \includegraphics[width=2.2cm]{fs/2.7515931129455566.jpg} & 
    \includegraphics[width=2.2cm]{morervl2/Two girls pose for a photo holding doughnuts..jpg}
    &\includegraphics[width=2.2cm]{morervl2/Two little girls are dressed in uniform preparing for the day.jpg}
    & \includegraphics[width=2.1cm]{morervl2/Two little girls sitting on the side of a white boat..jpg} \\
    \includegraphics[width=2.2cm]{morervl2/Two men next to each other holding wine glasses..jpg} 
    & \includegraphics[width=2.2cm]{morervl2/Two men smiling and taking a picture together..jpg}&
    \includegraphics[width=2.2cm]{morervl2/Two men work in the kitchen of a restaurant..jpg}
    & \includegraphics[width=2.2cm]{morervl2/Two teenage girls paying a video game together..jpg}
    & \includegraphics[width=2.2cm]{morervl2/Two women handing out cake on plates on a dining table..jpg}   \\
    \includegraphics[width=2.1cm]{morervl2/Woman in party dress sitting in beauty shop with hair dryer..jpg}&
    \includegraphics[width=2.1cm]{morervl2/A young male sitting on a park bench typing on a laptop..jpg} & 
    \includegraphics[width=2.1cm]{morervl2/A young man and woman holding their heads close to each other..jpg}
    &\includegraphics[width=2.1cm]{morervl2/A young man is working behind a counter..jpg}
    & \includegraphics[width=2.1cm]{morervl2/A young man sitting on a park bench next to a playground..jpg} \\
    \bottomrule
  \end{tabular}
  \caption{More images with faces generated by our fine-tuned model based on RV5.1 and $\ell_2$ distance measure.
  We omit the prompts for simplicity and focus on the face quality.}
  \label{fig:more results}
\end{figure}
\begin{figure}[ht]
  \centering
  \begin{tabular}{ccccc}
    \toprule  
    \includegraphics[width=2.1cm]{fs/-6.455127239227295.jpg} & \includegraphics[width=2.1cm]{fs/-5.948948383331299.jpg} & \includegraphics[width=2.1cm]{fs/-5.516416072845459.jpg} & \includegraphics[width=2.1cm]{fs/-4.702112674713135.jpg} & \includegraphics[width=2.1cm]{fs/-4.010016918182373.jpg} \\
     \midrule  
    -6.45 & -5.94 & -5.51 & -4.70 & -4.01 \\
    \midrule[1.3pt]
    \includegraphics[width=2.1cm]{fs/-3.770331621170044.jpg} & \includegraphics[width=2.1cm]{fs/-2.4210994243621826.jpg} & \includegraphics[width=2.1cm]{fs/-2.273516893386841.jpg} & \includegraphics[width=2.1cm]{fs/-1.6034494638442993.jpg} & \includegraphics[width=2.1cm]{fs/-1.342659831047058.jpg} \\
     \midrule  
    -3.77 & -2.42 & -2.27 & -1.60 & -1.34 \\
    \midrule[1.3pt]
    \includegraphics[width=2.1cm]{fs/-0.40085723996162415.jpg} & \includegraphics[width=2.1cm]{fs/0.07046686112880707.jpg} & 
    \includegraphics[width=2.1cm]{fs/0.7840189933776855.jpg} &\includegraphics[width=2.1cm]{fs/1.0660821199417114.jpg} & \includegraphics[width=2.1cm]{fs/1.8029941320419312.jpg} \\
     \midrule  
    -0.40 & 0.07 & 0.78 & 1.06 & 1.80 \\
    \midrule[1.3pt]
    \includegraphics[width=2.1cm]{fs/2.3492677211761475.jpg} & \includegraphics[width=2.1cm]{fs/2.7515931129455566.jpg} & 
    \includegraphics[width=2.1cm]{fs/3.0022552013397217.jpg} &\includegraphics[width=2.1cm]{fs/3.3463327884674072.jpg} & \includegraphics[width=2.1cm]{fs/4.140614032745361.jpg} \\
     \midrule  
    2.34 & 2.75 & 3.00 & 3.34 & 4.14 \\
    \midrule[1.3pt]
    \includegraphics[width=2.1cm]{fs/4.18411111831665.jpg} & 
    \includegraphics[width=2.1cm]{fs/4.255679130554199.jpg}&
    \includegraphics[width=2.1cm]{fs/4.687222957611084.jpg} & \includegraphics[width=2.1cm]{fs/4.944357395172119.jpg} & 
    \includegraphics[width=2.1cm]{fs/5.063962936401367.jpg}   \\
     \midrule  
    4.18 & 4.25 & 4.68 & 4.94 & 5.06 \\
    \midrule[1.3pt]
    \includegraphics[width=2.1cm]{fs/5.400645732879639.jpg}&
    \includegraphics[width=2.1cm]{fs/5.917778015136719.jpg} & 
    \includegraphics[width=2.1cm]{fs/6.016404628753662.jpg} &\includegraphics[width=2.1cm]{fs/6.1587958335876465.jpg} & \includegraphics[width=2.1cm]{fs/6.5052924156188965.jpg} \\
     \midrule  
    5.40 & 5.91 & 6.01 & 6.15 & 6.50 \\
    \bottomrule
  \end{tabular}
  \caption{More random face images and the corresponding \textit{Face Score} in increasing order.
  We can observe a trend that higher scores indicate better quality of facial generation. }
  \label{fig:more fs}
\end{figure}

\begin{figure}[t]
    \begin{tabular}{ccc|ccc}
        \textit{Score 1} & \textit{Score 2} & \textit{Score 3} & \textit{Score 1} & \textit{Score 2} & \textit{Score 3}\\
        \includegraphics[width=2cm]{evaluation/011.jpg}
        & \includegraphics[width=2cm]{evaluation/013.jpg}
        & \includegraphics[width=2cm]{evaluation/012.jpg} 
        & \includegraphics[width=2cm]{evaluation/021.jpg}
        & \includegraphics[width=2cm]{evaluation/022.jpg}
        & \includegraphics[width=2cm]{evaluation/023.jpg} \\
        \multicolumn{3}{c}{\multirow{2}{*}{\scriptsize \textit {\makecell{A young child standing in front of \\a table with plates of food}}
        }}
        &\multicolumn{3}{c}{\multirow{2}{*}{\scriptsize \textit {\makecell{A baby girl is holding a pink brush \\as she scratches her head}}
        }}\\[8pt]
        \includegraphics[width=2cm]{evaluation/031.jpg}
        & \includegraphics[width=2cm]{evaluation/032.jpg}
        & \includegraphics[width=2cm]{evaluation/033.jpg} 
        &\includegraphics[width=2cm]{evaluation/041.jpg}
        & \includegraphics[width=2cm]{evaluation/042.jpg}
        & \includegraphics[width=2cm]{evaluation/043.jpg} \\
        \multicolumn{3}{c}{\multirow{2}{*}{\scriptsize \textit {\makecell{A beautiful woman standing on the side of \\a rad next to a street}}
        }}
        &\multicolumn{3}{c}{\multirow{2}{*}{\scriptsize \textit {\makecell{A boy doing a skateboard trick on a street}}
        }}\\[8pt]
        \includegraphics[width=2cm]{evaluation/051.jpg}
        & \includegraphics[width=2cm]{evaluation/052.jpg}
        & \includegraphics[width=2cm]{evaluation/053.jpg} 
        &\includegraphics[width=2cm]{evaluation/061.jpg}
        & \includegraphics[width=2cm]{evaluation/062.jpg}
        & \includegraphics[width=2cm]{evaluation/063.jpg} \\
        \multicolumn{3}{c}{\multirow{2}{*}{\scriptsize \textit {\makecell{A boy dressed in a baseball uniform \\standing in a field}}
        }}
        &\multicolumn{3}{c}{\multirow{2}{*}{\scriptsize \textit {\makecell{A boy playing tennis on a \\blue and green tennis court}}
        }}\\[8pt]
        \includegraphics[width=2cm]{evaluation/071.jpg}
        & \includegraphics[width=2cm]{evaluation/072.jpg}
        & \includegraphics[width=2cm]{evaluation/073.jpg} 
        &\includegraphics[width=2cm]{evaluation/081.jpg}
        & \includegraphics[width=2cm]{evaluation/082.jpg}
        & \includegraphics[width=2cm]{evaluation/083.jpg} \\
        \multicolumn{3}{c}{\multirow{2}{*}{\scriptsize \textit {\makecell{A boy standing in the grass with a frisbee}}
        }}
        &\multicolumn{3}{c}{\multirow{2}{*}{\scriptsize \textit {\makecell{A boy with a blue shirt and jean pants \\doing a trick with his skateboard}}
        }}\\[8pt]
        \includegraphics[width=2cm]{evaluation/091.jpg}
        & \includegraphics[width=2cm]{evaluation/093.jpg}
        & \includegraphics[width=2cm]{evaluation/092.jpg} 
        &\includegraphics[width=2cm]{evaluation/101.jpg}
        & \includegraphics[width=2cm]{evaluation/102.jpg}
        & \includegraphics[width=2cm]{evaluation/103.jpg} \\
        \multicolumn{3}{c}{\multirow{2}{*}{\scriptsize \textit {\makecell{A child in a hat playing on a laptop computer}}
        }}
        &\multicolumn{3}{c}{\multirow{2}{*}{\scriptsize \textit {\makecell{A child riding a skateboard on a city street}}
        }}\\[8pt]       
        \includegraphics[width=2cm]{evaluation/111.jpg}
        & \includegraphics[width=2cm]{evaluation/112.jpg}
        & \includegraphics[width=2cm]{evaluation/113.jpg} 
        &\includegraphics[width=2cm]{evaluation/121.jpg}
        & \includegraphics[width=2cm]{evaluation/123.jpg}
        & \includegraphics[width=2cm]{evaluation/122.jpg} \\
        \multicolumn{3}{c}{\multirow{2}{*}{\scriptsize \textit {\makecell{A girl sitting on a bench \\in front of a stone wall}}
        }}
        &\multicolumn{3}{c}{\multirow{2}{*}{\scriptsize \textit {\makecell{A kid sitting on a bed with a remote}}
        }}\\[8pt] 
    \end{tabular}
    \caption{More examples of the human-annotated triplet. The image with higher face quality is assigned a higher score. 
    We can see despite RV5.1 having a smaller model size and fewer model parameters, its facial quality is comparable to that of SDXL. 
    }
    \label{fig:evaluation dataset}
\end{figure}

\begin{figure}[t]
    \centering
    \begin{tabular}{c|c}
        \includegraphics[width=5.5cm]{training_ds/children raising their hands in class.jpg}
        & \includegraphics[width=5.5cm]{training_ds/man in camouflage uniform.jpg} \\
         \scriptsize\textit{children raising their hands in class} & 
         \scriptsize\textit{man in camouflage uniform} \\
         \includegraphics[width=5.5cm]{training_ds/woman flying on a rocket pop art retro vector illustration.jpg}
        & \includegraphics[width=5.5cm]{training_ds/woman in white shirt and blue pants standing in front of a church.jpg} \\
        \multirow{2}{*}{\scriptsize \textit {\makecell{woman flying on a rocket \\pop art retro vector illustration}\vspace{15pt}}}
         & \multirow{2}{*}{\scriptsize \textit {\makecell{woman in white shirt and blue pants \\standing in front of a church}\vspace{15pt}}}\\[5pt]
         \includegraphics[width=5.5cm]{training_ds/man doing yoga in front of waterfall.jpg}
        & \includegraphics[width=5.5cm]{training_ds/man of steel hd wallpaper.jpg} \\
        \multirow{2}{*}{\scriptsize \textit {\makecell{man doing yoga in front of waterfall}\vspace{15pt}}}
         & \multirow{2}{*}{\scriptsize \textit {\makecell{man of steel hd wallpaper}\vspace{15pt}}}\\[5pt]
         \includegraphics[width=5.5cm]{training_ds/man of steel trailer.jpg}
        & \includegraphics[width=5.5cm]{training_ds/woman dancing on the beach.jpg} \\
        \multirow{2}{*}{\scriptsize \textit {\makecell{man of steel trailer}\vspace{15pt}}}
         & \multirow{2}{*}{\scriptsize \textit {\makecell{woman dancing on the beach}\vspace{15pt}}}\\[5pt]
         \includegraphics[width=5.5cm]{training_ds/man on a scooter.jpg}
        & \includegraphics[width=5.5cm]{training_ds/woman pushing shopping cart with christmas gifts.jpg} \\
        \multirow{2}{*}{\scriptsize \textit {\makecell{man on a scooter}\vspace{15pt}}}
         & \multirow{2}{*}{\scriptsize \textit {\makecell{woman pushing shopping cart \\with Christmas gifts}\vspace{15pt}}}\\[5pt]
          \includegraphics[width=5.5cm]{training_ds/woman with broken car and tire on road.jpg}
         \pagebreak
        & \includegraphics[width=5.5cm]{training_ds/man sitting on a scooter.jpg} \\
        \multirow{2}{*}{\scriptsize \textit {\makecell{woman with broken car and tire on road}\vspace{15pt}}}
         & \multirow{2}{*}{\scriptsize \textit {\makecell{man sitting on a scooter}\vspace{15pt}}}\\[5pt]
    \end{tabular}
\end{figure}

\begin{figure}[htbp]
    \centering
    \begin{tabular}{c|c}
    \includegraphics[width=5.5cm]{training_ds/woman in jeans and trench coat walking on the street.jpg}
        & \includegraphics[width=5.5cm]{training_ds/woman standing with recycling materials and recycling sign.jpg} \\
        \multirow{2}{*}{\scriptsize \textit {\makecell{woman in jeans and trench coat\\ walking on the street}\vspace{15pt}}}
         & \multirow{2}{*}{\scriptsize \textit {\makecell{woman standing with \\recycling materials and recycling sign}\vspace{15pt}}}\\[5pt]
          \includegraphics[width=5.5cm]{training_ds/man on a scooter.jpg}
        & \includegraphics[width=5.5cm]{training_ds/woman pushing shopping cart with christmas gifts.jpg} \\
        \multirow{2}{*}{\scriptsize \textit {\makecell{man on a scooter}\vspace{15pt}}}
         & \multirow{2}{*}{\scriptsize \textit {\makecell{woman pushing shopping cart \\with Christmas gifts}\vspace{15pt}}}\\[5pt]
          \includegraphics[width=5.5cm]{training_ds/woman in yellow dress standing on the beach holding a basket.jpg}
        & \includegraphics[width=5.5cm]{training_ds/woman in yellow sweater holding yellow suitcase and passport.jpg} \\
        \multirow{2}{*}{\scriptsize \textit {\makecell{woman in yellow dress standing \\on the beach holding a basket}\vspace{15pt}}}
         & \multirow{2}{*}{\scriptsize \textit {\makecell{woman in yellow sweater holding \\yellow suitcase and passport}\vspace{15pt}}}\\[5pt]
          \includegraphics[width=5.5cm]{training_ds/woman with arms up on top of car with ocean in background.jpg}
        & \includegraphics[width=5.5cm]{training_ds/woman wearing white shirt dress with red and white embroidery.jpg} \\
        \multirow{2}{*}{\scriptsize \textit {\makecell{woman with arms up on top of car \\with ocean in background}\vspace{15pt}}}
         & \multirow{2}{*}{\scriptsize \textit {\makecell{woman wearing white shirt dress \\with red and white embroidery}\vspace{15pt}}}\\[5pt]
          \includegraphics[width=5.5cm]{training_ds/man sitting on floor with boxes and cleaning supplies.jpg}
        & \includegraphics[width=5.5cm]{training_ds/man pouring wine.jpg} \\
        \multirow{2}{*}{\scriptsize \textit {\makecell{man sitting on floor with boxes \\and cleaning supplies}\vspace{15pt}}}
         & \multirow{2}{*}{\scriptsize \textit {\makecell{man pouring wine}\vspace{15pt}}}\\[5pt]
    \end{tabular}
    \caption{More examples of face pairs. We leverage the inpainting pipeline and control the noise factor for a degraded version, thereby forming a (good, bad) face pair.}
    \label{fig: ds}
\end{figure}

%
%